\makeatletter
\newcommand{\myparagraph}[1]{\paragraph{#1}\mbox{}\\}
\makeatother

\documentclass[a4paper, 11pt, oneside]{Thesis}  
\graphicspath{Figures/}  

\usepackage{multirow}
\usepackage{graphicx}
\usepackage{wrapfig}
\usepackage{graphbox}
\usepackage{subcaption}
\usepackage{amsmath}

\usepackage{verbatim}  
\usepackage{vector}  
\usepackage{natbib}
\begin{document}
\frontmatter      

\UNIVERSITY{{THE UNIVERSITY OF MELBOURNE }}    
%
\department{{School of Computing and Information Systems}}
\school{{Melbourne School of Engineering}}

%
\title  {Language Model Evaluation in Open-ended Text Generation}
\authors  {\texorpdfstring
            {\href{your web site or email address}{An Nguyen\\ \small Student ID: 1098402 }}
            {Author Name}
            }
\addresses  {\groupname\\\deptname\\\univname}  
\date       {\today}
\subject    {}

\maketitle

\setstretch{1.3}  

\fancyhead{}  
\rhead{\thepage}  
\lhead{}  

\pagestyle{fancy}  


%
%
\Declaration{

\addtocontents{toc}{\vspace{1em}}  

I confirm that:

\begin{itemize} 
\item[\tiny{$\blacksquare$}] This thesis does not incorporate without acknowledgement any material previously submitted for a degree or diploma in any university; and that to the
best of my knowledge and belief it does not contain any material previously
published or written by another person where due reference is not made in the
text;
 
\item[\tiny{$\blacksquare$}] Where necessary I have received clearance for this research from the University’s Ethics Committee and have submitted all required data to the School;

\item[\tiny{$\blacksquare$}] The thesis is 12,714 words in length (excluding text in images, table, bibliographies and appendices).
\\
\end{itemize}

Signed: \textbf{An Nguyen}\\
\rule[1em]{25em}{0.5pt}  
 
Date: \textbf{06/01/2021}\\
\rule[1em]{25em}{0.5pt}  
}
\clearpage  

%
%
\Preface{

\addtocontents{toc}{\vspace{1em}}  

Although current state-of-the-art language models have achieved impressive results in numerous natural language processing tasks, still they could not solve the problem of producing repetitive, dull and sometimes inconsistent text in open-ended text generation. Studies often attribute this problem to the maximum likelihood training objective, and propose alternative approaches by using stochastic decoding methods or altering the training objective. However, there is still a lack of consistent evaluation metrics to directly compare the efficacy of these solutions. In this work, we study different evaluation metrics that have been proposed to evaluate quality, diversity and consistency of machine-generated text. From there, we propose an practical pipeline to evaluate language models in open-ended generation task, and research on how to improve the model's performance in all dimensions by leveraging different auxiliary training objectives.
}
\clearpage  

\setstretch{1.3}  
\acknowledgements{
\addtocontents{toc}{\vspace{1em}}  

First and foremost, I would like to express my sincere appreciation to my supervisor, Dr. Jey Han Lau, who has constantly given me thoughtful guidance and valuable feedback throughout my thesis. I’m incredibly fortunate to have you as my supervisor - thank you so much for all of the encouragement and support for me during the middle of this global pandemic.

I’m eternally grateful for my girlfriend, Ngan, who has always been beside me and supported me during the most difficult time.

Finally, I want to express my profound gratitude to my parents and sister, who have always loved and supported me unconditionally.

}
\clearpage  

\pagestyle{fancy}  

\lhead{\emph{Contents}}  
\tableofcontents  

\lhead{\emph{List of Figures}}  
\listoffigures  

\lhead{\emph{List of Tables}}  
\listoftables  






\mainmatter	  
\pagestyle{fancy}  


\lhead{\emph{Introduction}}
\chapter{Introduction}

Language modeling is the task of calculating the probability distribution over word sequences, with a goal to generate more \textit{fluent} text in a certain language, where having \textit{higher probability} is considered \textit{fluent}. In recent years, neural language modeling has become the powerhouse for an array of natural language processing tasks, from machine translation \citep{bahdanau2014neural, luong2015effective}, summarization \citep{zhang2019pegasus} to story generation \citep{fan2018hierarchical}. The introduction of the Transformers architecture \citep{vaswani2017attention} has allowed language model to be trained on even more massive text datasets in significantly less time thanks to its ability to parallelize computation. Today, large pre-trained language model like GPT-2 \citep{radford2019language}, or the latest GPT-3 \citep{brown2020language} with 175 billion parameters have achieved state-of-the-art results in numerous tasks in zero-shot and few-shot setting.

Despite of their superiority in multiple NLP tasks, these language models are still falling short in open-ended text generation task such as story generation and dialogue modelling, where the model is required to produce long continuation of text. Based on empirical observation with standard deterministic decoding method, the text is often found to be dull, repetitive \citep{holtzman2019curious, welleck2019neural, shao2017generating, fan2018hierarchical}, and sometimes logically inconsistent and factually incorrect although being fluent and coherent \citep{li2019dont, welleck2019dialogue, hayashi2019latent, petroni2019language}.

There exists a number of solutions to this problem, such as using stochastic decoding methods when generate continuations to keep the model from repeating itself \citep{fan2018hierarchical, holtzman2019curious}, or altering the training objective to penalize the model from being repetitive \citep{welleck2019neural, bengio2015scheduled}. However, to the best of our knowledge, there has not been any works that \textit{quantitatively} compare the performance of these solutions, due to the lack of a consistent evaluation metric. Therefore, it is hard to decide which solution is better than the others.

Traditionally, language models have been evaluated based on perplexity, which concerns with the probability of a sentence being produced by the model. We also have a number of other metrics for different tasks, such as BLUE for machine translation \citep{papineni2002bleu} or ROUGE for text summarization \citep{lin2004rouge}, which essentially compare the similarity between human and machine generated text. This works when we want to judge the \textit{quality} of the generated text - we want our models to produce natural, human-like and grammatically correct sentences.

However, with open-ended generation task such as story telling or dialogue generation, not only we want our model to produce high quality text but also to be \textit{creative} and \textit{diverse}. For example, given the prompt \textit{``Once upon a time"}, we expect our story generation model to generate a diverse range of continuations rather than repeating the most probable story again and again. This is where traditional metrics like perplexity can be problematic, since it places a high stress on the probability of individual words. In terms of creativity or diversity, we would probably prefer seeing \textit{``she takes the yacht to school"} than \textit{``she takes the bus to school"} in a story. However, these two sentences can have significantly different probabilities because the word \textit{``yacht"} is much less probable than \textit{``bus"}. Thus, the first sentence might result in much higher perplexity, which can be perceived as being lower in quality.

Another important aspect to open-ended text generation is commonsense reasoning, which we will refer to as \textit{consistency}. Since the models are expected to produce much longer text, they are more prone to generating illogical or factually incorrect sentences \citep{zhang2018personalizing, welleck-etal-2019-dialogue, li2019dont}. For example, when generating an answer to a customer in a dialogue, a chat bot can say that \textit{``our store remains open in the weekend"} then immediately say \textit{``our store will be closed on Saturday"}. Perplexity simply cannot capture the quality of being consistent or logical of a generative model.




\section{Objectives}

\subsection{Rethinking language models evaluation}

\begin{quote} 
\centering 
\textit{Question: What is the best metric to use when evaluating language models on open-ended text generation task in each dimension: quality, diversity and consistency?}
\end{quote}

It is important that when evaluating language models on open-ended generation task, we need to look at their performance in all three dimensions: quality, diversity and consistency. However, in each dimension, there exists a number of different metrics that have been proposed in the literature. For example, to evaluate the quality of machine-generated text, one can use either Corpus-BLEU \citep{yu2017seqgan} or forward perplexity \citep{zhao2018adversarially}. If two studies make use of different metrics to evaluate their models performance, then it becomes impossible to directly make comparison between the two.

In this work, we study different evaluation metrics that have been proposed to evaluate quality, diversity and consistency of machine-generated text, and aim to find the best metric to use in each dimension.

\subsection{Stochastic Decoding or New Training Objectives?}

\begin{quote} 
\centering 
\textit{Question: Which technique can lead to better performance of language models in open-ended text generation task, stochastic decoding methods or tweaking the training objective?}
\end{quote}

Although stochastic decoding methods significantly reduced the repetition issue thanks to randomization \citep{holtzman2019curious, fan2018hierarchical}, they do not solve the underlying problem with maximum likelihood training. Because of this, altering the training objective might sound like a more viable approach. Many works has been carried out in this regards, such as scheduled sampling \citep{bengio2015scheduled}, Generative Adversarial Nets \citep{goodfellow2014generative, yu2017seqgan, guo2017long}, or most recently unlikelihood training \citep{welleck2019neural, li2019dont}. 

However, since there has not been any direct comparison between the two techniques, it is hard to decide which one is superior. We argue that if both lead to the same performance, we would be better off using stochastic decoding methods since it is cheaper in terms of training. Using the new evaluation pipeline that we have proposed in this work, we aim to find out which one is the superior technique to use for language models in open-ended text generation task: stochastic decoding methods or tweaking the language model training objective.



\subsection{Multi-task learning}

\begin{quote} 
\centering 
\textit{Quesiton: Can multi-task learning help language models to generate better quality, more diverse and more consistent text?}
\end{quote}

With the success of BERT in a plethora of different NLP tasks \citep{devlin-etal-2019-bert}, there has been a surge of interest in multi-task learning for language models. Besides traditional maximum likelihood estimate objective, many different auxiliary training objectives have been introduced, such as masked language model \citep{devlin-etal-2019-bert, yang2019xlnet}, next sentence prediction \citep{devlin-etal-2019-bert}, or word/sentence order prediction \citep{wang2019structbert}. All of these additional objectives share a common goal: to make the language model become better at \textit{understanding} the language. In this work, we are curious to find out if multi-task learning can help languages model to become better on a specific task - open-ended text generation. We only focus on auxiliary training objectives in an unsupervised setting, i.e. where training labels can be obtained automatically. 

\section{Thesis Overview}

\subsection{Chapter 2}

In Chapter 2, we first provide necessary backgrounds to understand language models and their evolution, from simple probabilistic n-gram models to state-of-the-art Transformer-based models. We then illustrate the neural text degeneration problem, and compare different solutions that have been proposed in the literature. We end this chapter by looking at different evaluation metrics for language models on open-ended text generation task in all dimensions - quality, diversity and consistency.

\subsection{Chapter 3}

In Chapter 3, we present our experiment to compare different evaluation metrics for language models on open-ended text generation task. We decide what is the best metric to use for each dimension, then propose a evaluation pipeline using these metrics and apply it to compare the performance between different stochastic decoding methods and unlikelihood training. We also carry out an experiment to see how the choice of domain for training corpus can effect the consistency of continuations.

\subsection{Chapter 4}

In Chapter 4, we give an overview of multi-task learning, then present our experiment in which we fine-tune language models with different auxiliary training objectives. Using the proposed pipeline from Chapter 3, we evaluate the efficacy of these auxiliary training objectives in open-ended text generation task.

\subsection{Chapter 5}

In Chapter 5, we give a conclusion of our project and its contribution, and provide several suggestions for future work. 

\lhead{\emph{Background}}
\chapter{Background}

\section{Language Models}

In recent years, language models have become the powerhouse for an array of natural language processing (NLP) tasks, from machine translation \citep{bahdanau2014neural, luong2015effective}, summarization \citep{zhang2019pegasus}, story generation \citep{fan2018hierarchical} to dialogue generation \citep{li2019dont, zhang2018personalizing, welleck2019dialogue}. So what exactly is a language model?

To formalize, for a text that contain $m$ tokens $\{w_1...w_m\}$, a language model is used to compute the probability of that text: 

$$p(w_1, w_2, ..., w_m), w \in V$$

where $V$ is the set of vocabulary. The goal of computing this probability is to produce more fluent texts, which should have much higher probability than odd ones. For example, if we want to translate the following sentence from Vietnamese to English: \textit{``Tôi thích xe màu cam và màu xanh"}, the literal word-by-word translation would be \textit{``I like car orange and blue"}. The language model should give this translation a lower probability than a natural translation like \textit{``I like orange and blue cars"}.

To generate text, a language model can produce the next word given a context by using the conditional probability of the next word on the previous words:

$$p(w_k | w_1, ..., w_{k-1}), w \in V$$

In the rest of this section, we explore the history of language models and reveal how we have reached the states-of-the-art models at present, which stand behind recent remarkable achievements in multiple NLP tasks.

\subsection{n-gram}

To compute the probability of a sequence of text, we can refactor the joint probability using the chained rule:

$$p(w_1, w_2, ..., w_m) = p(w_1) \cdot p(w_2|w_1) \cdot p(w_3|w_1,w_2) \cdot ... \cdot p(w_m|w_1,...,w_{m-1})$$

This equation suggests that at every step, we have to calculate the probability of word given all of its \textit{predecessors}. This can be done by simply counting the number of occurrences of the combination with and without the last word in the whole universe. For example, for the sentence \textit{``I like orange and blue cars"}, the probability of the word \textit{``blue"} is equal to:

$$p(\textrm{blue}|\textrm{I like orange and}) = \frac{count(\textrm{I like orange and blue})}{count(\textrm{I like orange and})} $$

In practice, calculating probabilities by counting every existing word combinations is impossible, as we would never have a large enough dataset to produce a correct estimation, especially for long sequences. To make the calculation tractable, an n-gram language model simplifies this and calculate the probability of a word using only the $n-1$ previous words:

$$ p(w_1, w_2, ..., w_m) = \prod_{k=1}^{m} p(w_k|w_{k-n+1},...,w_{k-1})$$

For example, in a uni-gram (1-gram) language model, the probability of the sentence \textit{``I like orange and blue cars"} will be computed as:

$$p(\textrm{I}|\textrm{\textlangle s\textrangle}) \cdot p(\textrm{like}|\textrm{I}) \cdot p(\textrm{orange}|\textrm{like}) \cdot p(\textrm{and}|\textrm{orange}) \cdot p(\textrm{blue}|\textrm{and}) \cdot p(\textrm{cars}|\textrm{blue})$$

An obvious problem with n-gram language models is \textit{sparsity}, i.e. yielding zero probability for unknown word combinations. This is very likely to happen in practice, as language is always evolving and new combinations are created everyday. There exists a number of solutions to this problem, such as back-off \citep{kneser1995improved} or smoothing \citep{chen1999empirical}. A general idea of these solutions is to give the unknown combination a tiny amount of the probability mass, so we would not encounter any zero values in our calculation.

Another problem with n-gram language models is that they are very limited in modeling long-range dependency between words in a sentence. For example, the sentence \textit{``The cat at the end of the street comes here everyday"} requires at least an 8-gram language model in order to know that \textit{``comes"} should be a singular verb because of the singular noun \textit{``cat"}. Increasing $n$ is not a viable solution, as n-gram language models can be computationally expensive when $n$ is large. In the last example, an 8-gram language model with the vocabulary size of 100,000 words can have up to $10^{40}$ possible sequences.

Lastly, n-gram language models are incapable of informing us about the linguistic and semantic information of the language, as everything in n-gram language models are just plain statistic. Given the sentence \textit{``a man is eating an apple"}, n-gram language models have no way to recognize that it is actually very similar to the sentence \textit{``a woman is eating an orange"} in term of semantic. This is because in n-gram language models we are only looking at the surface forms of words, thus in the view of an n-gram language model, the difference between the word \textit{``man"} and \textit{``woman"} is the same as the difference between \textit{``man"} and \textit{``bread"}. Therefore, n-gram language models do not have the ability to generalize their knowledge to sequences that they have not encountered during training.

\subsection{Feed-forward neural language model}

\subsubsection{Feed-forward neural network}
\begin{figure}[h]
\centering
\includegraphics[width=12cm]{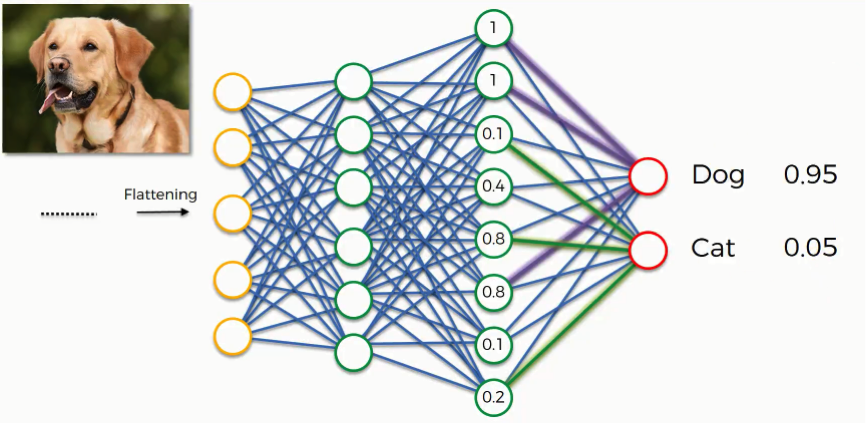}
\caption{Example of a feed-forward neural network \citep{bengio2003neural}}
\label{DogCatNN}
\end{figure}

A neural network is a collection of computing units whose goals are to learn a mapping function between a set of inputs and desired outputs. For example, we might want to build a neural network to predict whether an image is a picture of a cat or a dog. In this case, the inputs can be the individual pixels of the image, and the outputs can be the probability of this image being a picture of a dog or a cat (Figure \ref{DogCatNN}). Each computation unit has an associated \textit{weight}, which can be learned with the help of a \textit{loss function} to penalize the network when it makes incorrect predictions. The end goal is to learn a set of weights that maximize \textit{the likelihood} of the training data.

This neural network in Figure \ref{DogCatNN} is called feed-forward because of its architecture: the computation proceed forward with no cycle. A neural network can have multiple hidden layers between the input layer and output layer, which is often known as a \textit{deep} neural network.

\subsubsection{Feed-forward neural language model}

\begin{figure}[h]
\centering
\includegraphics[width=12cm]{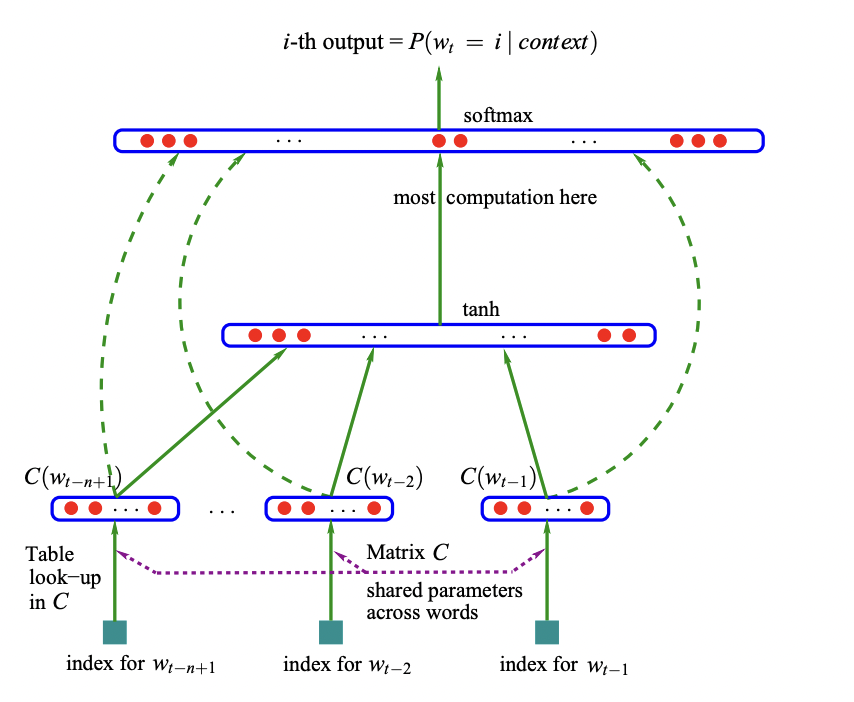}
\caption{Classic feed-forward neural network language model \citep{bengio2003neural}}
\label{BengioLanguageModel}
\end{figure}

The first feed-forward neural language model was proposed by \citet{bengio2003neural}, whose architecture is shown in Figure \ref{BengioLanguageModel}. Similar to a n-gram language model, the feed-forward neural language model uses the $n$ previous words as context to compute the probability of the next word in the sequence. What differs here is that each context word has a vector representation, which can be looked up in a table $C$. These vectors are concatenated and passed to a hidden layer, follow by a final softmax layer to obtain the probability distribution of the next word over the entire vocabulary.

This model has the same downside as n-gram language models, since it only has access to $n$ previous words to predict the next word. However, feed-forward neural language models do not require smoothing (technically the softmax layer never produce zero-values due to the exponential operation), and they can generalize much better over similar contexts. Recall the example from last section, where the n-gram model cannot know that the sentence \textit{``a man is eating an apple"} is actually very similar to the sentence \textit{``a woman is eating an orange"}. The feed-forward neural language model solves this problem by having access to a dense vector representation for each word, such that similar words are expected to have similar feature vectors. This way, when the model updates its parameters following a specific word, the changes will be carried over to similar words as well.

\subsubsection{Word embeddings}

\citet{bengio2003neural}'s language model has laid the foundation for what we now know as \textit{word embedding} - a real-valued vector used to represent words, often ranges between ten to few hundreds of dimensions. There exists a number of pre-trained word embeddings, among which the most notable ones are GloVe \citep{pennington2014glove} and word2vec \citep{mikolov2013efficient, mikolov2013distributed}. Compare to \citet{bengio2003neural}, GloVe and word2vec's training objectives are much simpler, allowing them to be trained on a much larger scale. Thus, using these pre-trained word embeddings directly on downstream tasks is much more effective than training an word embedding layer from scratch in terms of both training speed and task performance \citep{wang2020survey}.

\begin{figure}[h]
\centering
\includegraphics[width=12cm]{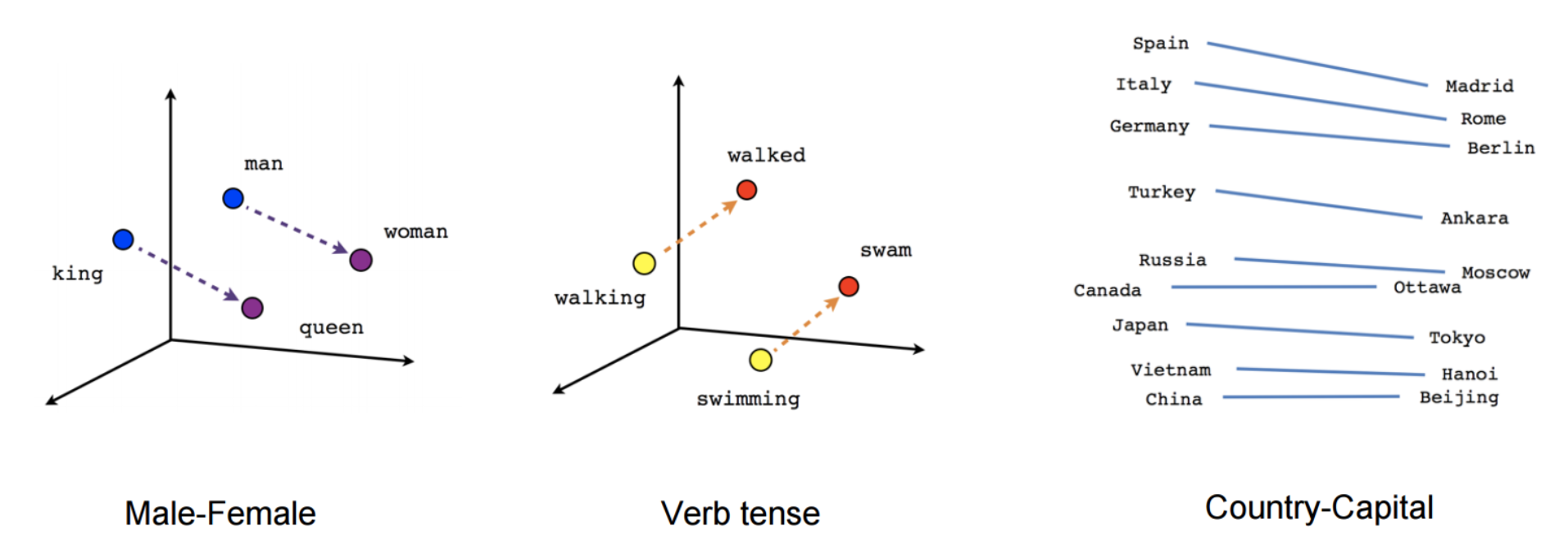}
\caption{Example of relations captured by word2vec word embeddings when being projected to a low dimension space \citep{mikolov2013efficient, mikolov2013distributed}. Image credit: \citet{ruder2018review}.}
\label{Word2VecRelations}
\end{figure}

Interestingly, word embeddings have been shown to be capable of capturing linguistic relationships between words (Figure \ref{Word2VecRelations}). This suggests that word embeddings can in fact represent the meaning of a word, which is a huge step towards natural language understanding.

\subsection{Recurrent neural language model}

\subsubsection{Recurrent neural network}

\begin{figure}[h]
\centering
\includegraphics[width=12cm]{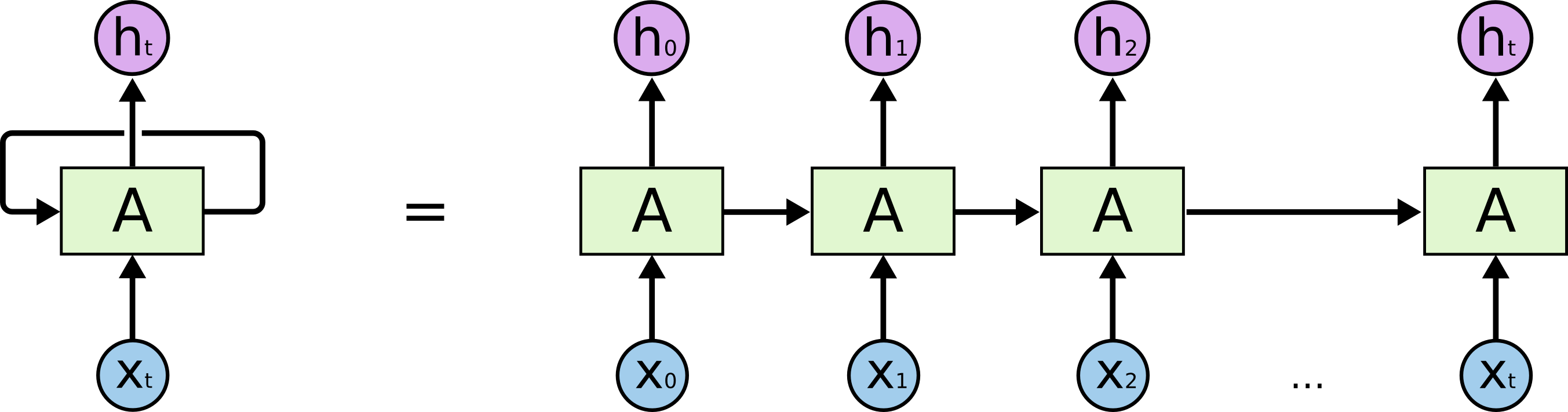}
\caption{Recurrent neural network \citep{colahRNN2015}}
\label{RNNModel}
\end{figure}

A recurrent neural network (RNN) is a special type of neural network that contain cycles in its structure. It is typically used for sequential data, such as audio signals, time series or languages. An RNN processes the input sequence one at a time, and maintains a state vector (known as \textit{the hidden state}) to store the information of previously processed contexts. An example of an RNN is given in Figure \ref{RNNModel}.

We refer to a specific point in the sequence as a \textit{time step}. At every time step $t$, the input $x_t$ is passed into the network. Using this input, along with the hidden state from the last time step $h_{t-1}$, the model computes the current hidden state $h_t$. This process is repeated until we hit the end of the sequence. Depends on the type of recurrent neural network, the model can choose to yield an output $y_t$ at every time step (one-to-one, e.g. part-of-speech tagging), yield a single output $y_T$ at the final step (many-to-one, e.g. text classification), or produce a different length $T'$ outputs $\{y_1,...,y_T'\}$ at the final step (many-to-many, e.g. machine translation).

\subsubsection{Recurrent neural language model}

\begin{figure}[h]
\centering
\includegraphics[width=12cm]{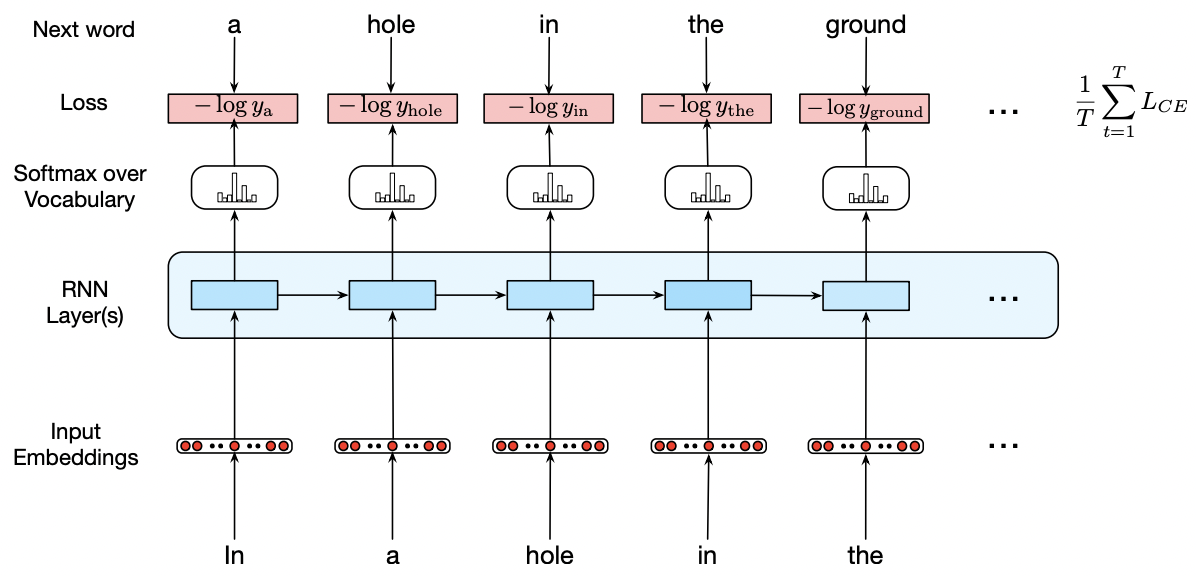}
\caption{Example of RNN language model \citep{jurafsky2009speech}}
\label{RNNLanguageModel}
\end{figure}

The use of RNN in language modeling was first introduced by \citet{mikolov2010recurrent}. Figure \ref{RNNLanguageModel} gives an example of how we can train a recurrent neural language model. With each sequence used for training, we can obtain the target labels by shifting the whole sequence one step to the left. At every time step, we pass the word embedding of the current word to the RNN layers, apply the softmax operation to obtain the probability distribution of the next word, then calculate the loss for back propagation using the target word. RNN language models have been shown to outperformed all state of the art n-gram models at the time, even when the n-gram models were given much more training data \citep{mikolov2010recurrent}.

Thanks to their recurrent structure, RNN language models do not rely on a fixed length context, but can process a sequence of arbitrary length. Every word has access to the latest hidden state, which carries the information of every other previous words in the sequence. In practice however, RNN still struggles to capture long-range dependencies, as information from early time steps can diminish at later ones, especially in long sequence - a problem known as \textit{vanishing gradients}. In addition, because of its sequential nature, RNN is much slower to trained compare to than feed-forward neural network.

\subsubsection{Long short-term memory}

\begin{figure}[t!]
    \centering
    \begin{subfigure}[t]{\textwidth}
        \centering
        \includegraphics[width=12cm]{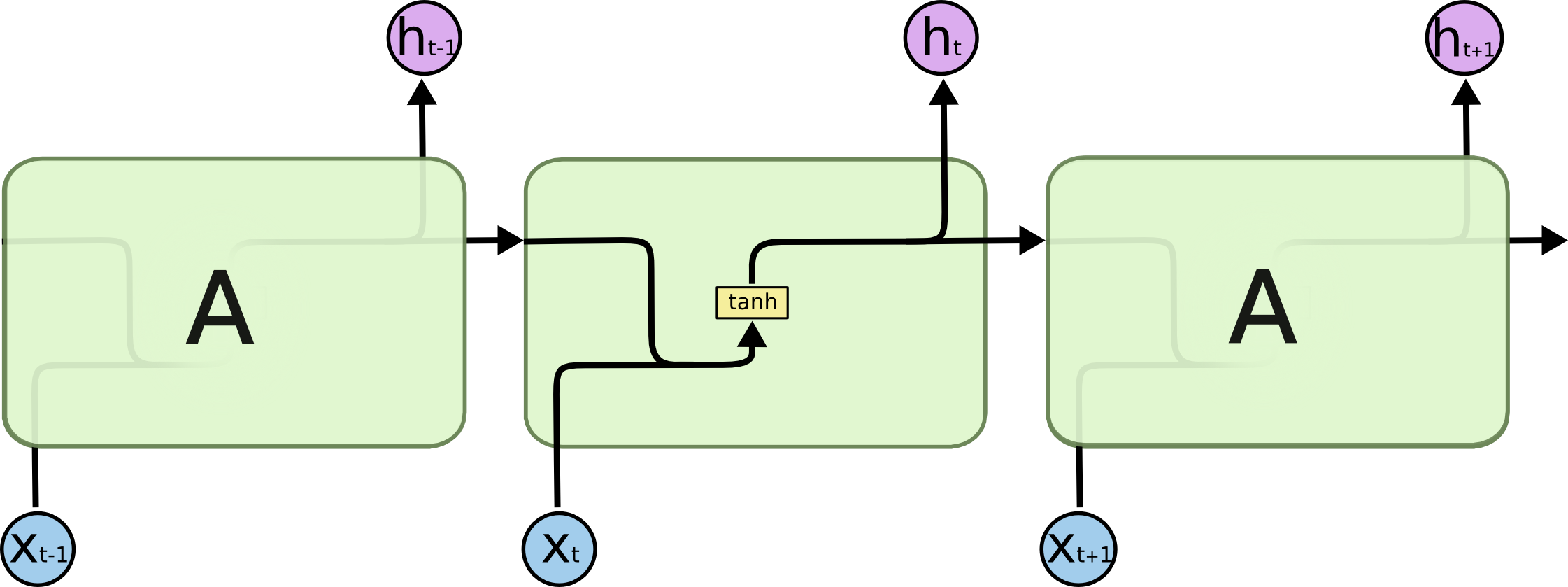}
        \caption{RNN}
    \end{subfigure}

    \begin{subfigure}[t]{\textwidth}
        \centering
        \includegraphics[width=12cm]{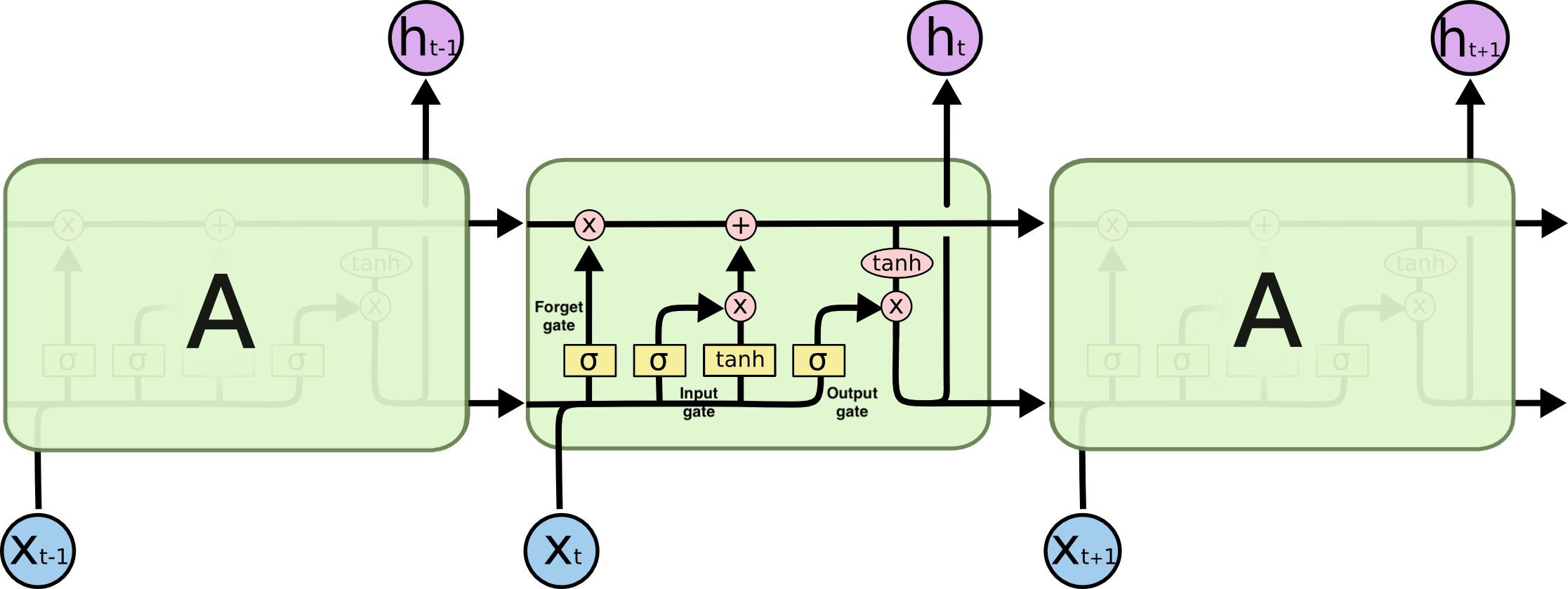}
        \caption{LSTM}
    \end{subfigure}

    \caption{Comparison between RNN and LSTM \citep{colahRNN2015}}
    \label{RNNvsLSTM}
\end{figure}

To solve the vanishing gradient problem, \citet{hochreiter1997long} proposed the use of the long short-term memory (LSTM) block to replace the traditional RNN block, with the addition of memory cells to preserve gradients throughout the sequence, and multiple gates to control access to these memory cells. A comparison between the traditional RNN block and LSTM block is given in Figure \ref{RNNvsLSTM}.

The three type of gates in the LSTM block are the input gate, forget gate and output gate. At a high level, the forget gate determines how much of the current memory cell's content should be forgotten, the input gate decides how much knowledge should be added to the new memory cell, and the output gate controls how much of the memory cell's content should be transferred to the next hidden state. For example, when processing the word \textit{``he"} in the sentence \textit{``The two cars that he loves"}, the forget gate can choose to let go of the information about the plural noun \textit{``cars"}, and the input gate can store this new information about the singular pronoun \textit{``he"} in order to correctly predict the singular verb \textit{``loves"}.

\subsubsection{Sequence-to-sequence}

\begin{figure}[h]
\centering
\includegraphics[width=12cm]{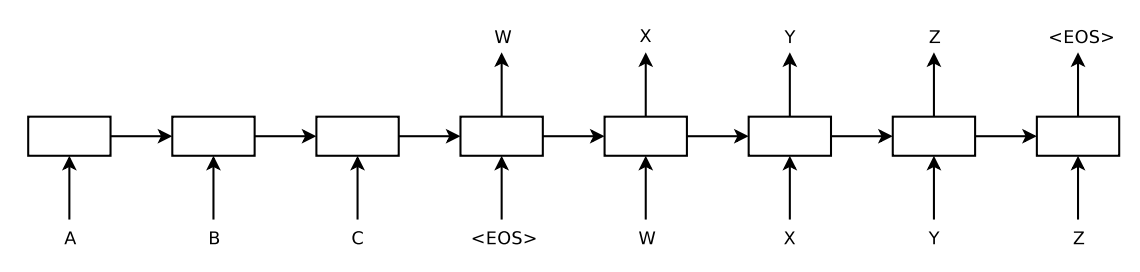}
\caption{Example of sequence-to-sequence (encoder-decoder) architecture, where the model encodes the sentence ABC and outputs the sentence WXYZ \citep{sutskever2014sequence}}
\label{Seq2Seq}
\end{figure}

Another variant of recurrent neural network language models is the sequence-to-sequence (also known as encoder-decoder) model, which deals with generation tasks where the length of the output sequence is different from the input sequence. The model was first introduced by \citet{sutskever2014sequence}, and it has been widely used in a variety of natural language processing tasks, such as neural machine translation or image captions generation. It works by first processing and compressing the input sequence into a fixed-sized vector using the encoder network, then uses that as the first hidden state to start decoding the output sequence using the decoder network. An example of this architecture is given in Figure \ref{Seq2Seq}.

This compression step might become a bottleneck, as the source sequence has to be squeezed into a fixed-size vector before passing to the decoder network. If we have a long source sequence, it becomes difficult for the model to transfer every information of the source sequence in the encoder network to the decoder network.

\subsection{Attention and Transformer}

\subsubsection{Attention mechanism}

Unlike the sequential process in seq2seq models, when translating a sentence from a language to another, human do not necessarily take just one single look at the source sentence and immediately translate it to the target language. What we tend to do is rather a iterative process, where we keep working back and forth between the source sentence and the target sentence to find the information needed \textit{at a specific time step} and ignore the rest.

\begin{figure}[h]
\centering
\includegraphics[width=6cm]{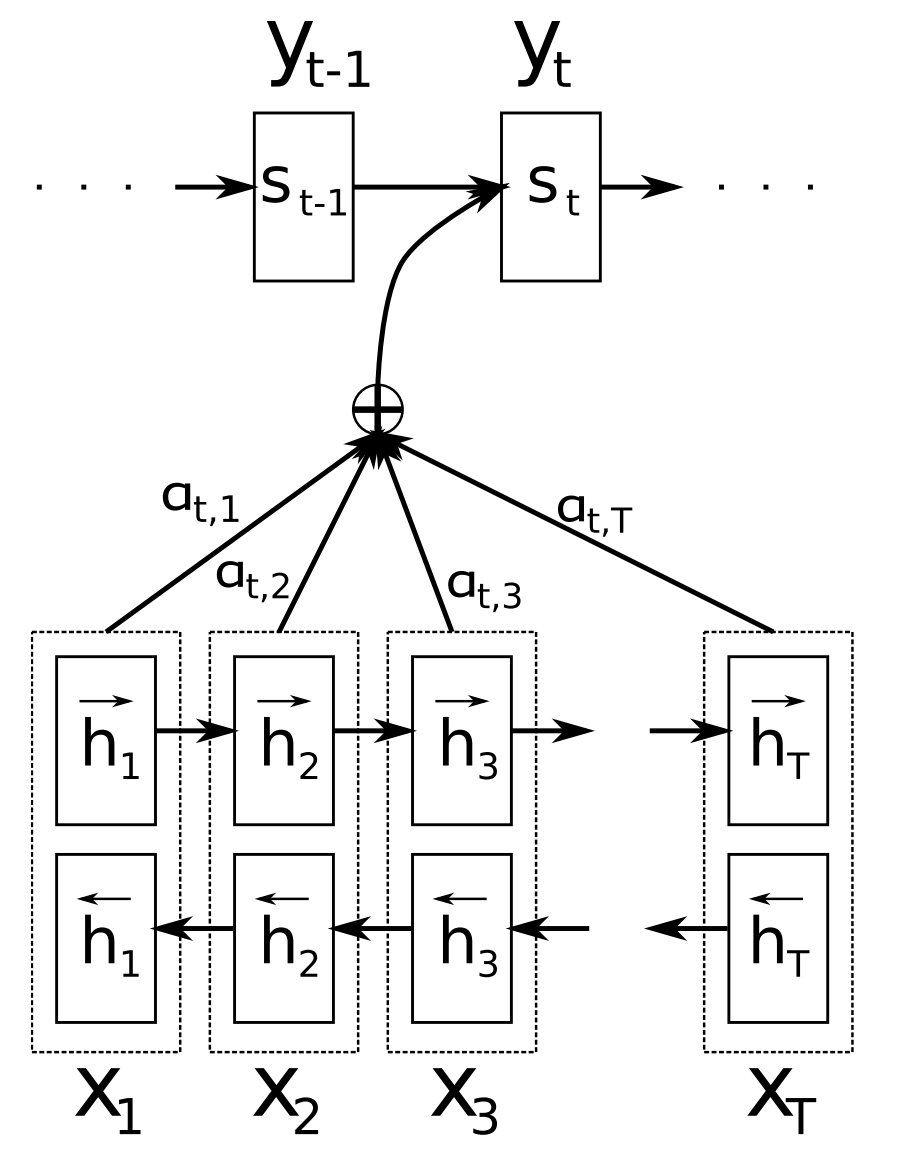}
\caption{Attention in machine translation \citep{bahdanau2014neural}}
\label{AttentionMT}
\end{figure}

This example in language translation is exactly the motivation for the attention mechanism, which was first proposed by \citet{bahdanau2014neural}. In this model, the attention mechanism allows the decoder network to look at \textit{every hidden state} from the encoder network to find all the information it needs. In addition to the use of the last hidden state $s_{t-1}$ and the last output $y_{t-1}$ to produce the next hidden state $s_t$ in the decoder network, Bahdanau et al. introduce a context vector $c_t$ - a weighted summation of all hidden states from the encoder network (Figure \ref{AttentionMT}):

$$s_t = f(s_{t-1}, y_{t-1}, c_t)$$
$$c_t = \sum_{i=1}^{T_x} \alpha_{ti} \cdot h_i$$

Attention has been successfully applied to a plethora of NLP tasks, such as machine translation \citep{bahdanau2014neural, luong2015effective}, text classification \citep{letarte2018importance, jain2019attention}, text summarization \citep{rush2015neural, wu2018word, zhang2019pegasus} or question answering \citep{sukhbaatar2015end, kim2017structured}. Despite the significant improvements in task performance, the slow computation time of seq2seq models remains an unsolved problem, preventing them to be trained on a larger scale.

\subsubsection{Transformer and self-attention}

\begin{figure}[h]
\centering
\includegraphics[width=8cm]{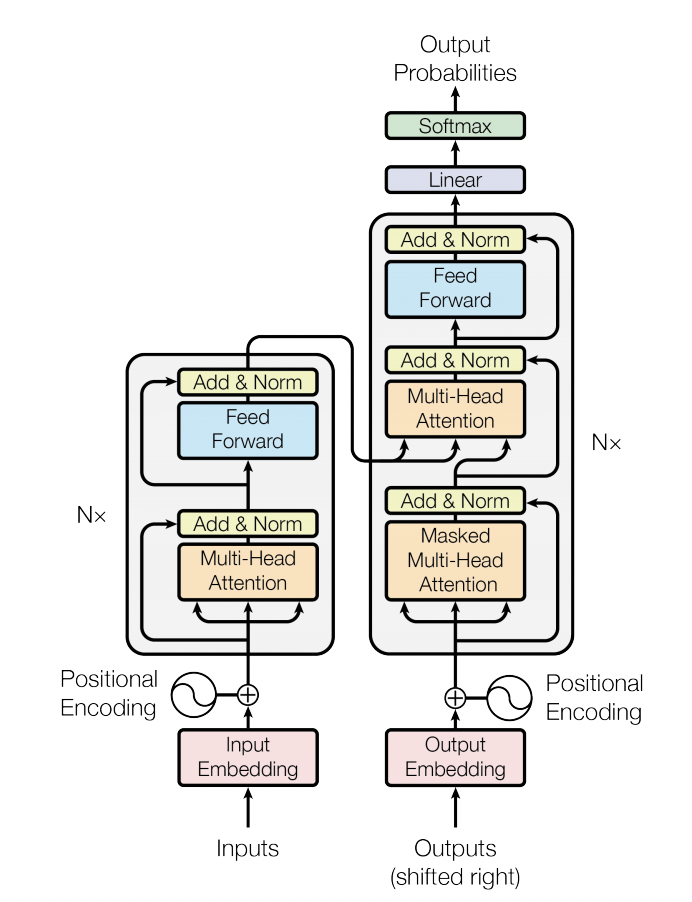}
\caption{Transformer architecture \citep{vaswani2017attention}}
\label{TransformerArchitecture}
\end{figure}

In the seminal paper \textit{Attention is all you need}, \citet{vaswani2017attention} has revolutionized the NLP field with the Transformer block (Figure \ref{TransformerArchitecture}), which completely eliminates the need for sequence processing in RNN and seq2seq models. The main innovation in the Transformer block comes from the multi-head attention layer.

\begin{figure}[t!]
    \centering
    \begin{subfigure}[t]{0.5\textwidth}
        \centering
        \includegraphics[width=6cm]{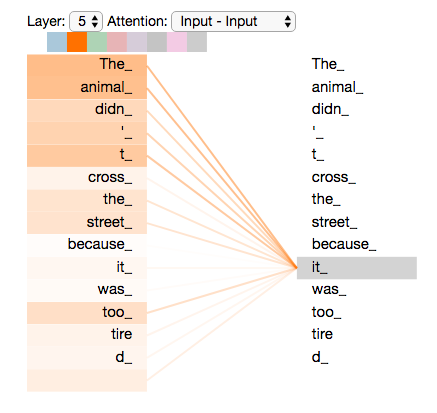}
        \caption{Single-head attention}
        \label{SingleHeadAttentionExample}
    \end{subfigure}%
    ~
    \begin{subfigure}[t]{0.5\textwidth}
        \centering
        \includegraphics[width=6cm]{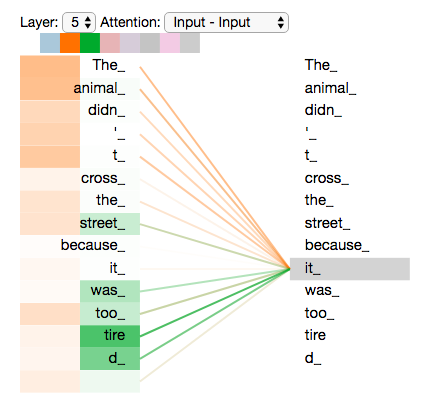}
        \caption{Multi-head attention}
        \label{MultiHeadAttentionExample}
    \end{subfigure}

    \caption{Example of self-attention. Image credit: \citet{alammar2018transformer}}
    \label{SelfAttentionExample}
\end{figure}

To understand multi-head attention, we first need to understand what self-attention is. Whereas RNN maintains the current hidden state by looking at the previous hidden state/output vector, self-attention allows the model to construct a hidden state at each time step by attending to \textit{every other input} in the sequence. Borrowing an example from \citet{alammar2018transformer}, where we want to process the following sentence \textit{``The animal didn't cross the street because it was too tired"} using the Transformer network. Figure \ref{SingleHeadAttentionExample} illustrates how the model construct the hidden state for the word \textit{``it"} by looking at other positions in the input sequence. Note that in this example, it gives most of the weights to the phrase \textit{``the animal"}, which is exactly what the word \textit{``it"} is referring to.


A word in a sentence can relate to other words in various ways, e.g. being a nominal subject for a word while being a direct object to another. This is exactly what multi-head attention can be used for: each \textit{head} in the multi-head attention layer is a self-attention layer with its own parameters, so it can learn about different relationships between the source word and the others simultaneously. Figure \ref{MultiHeadAttentionExample} gives an example of this phenomenon, where the word \textit{``it"} attends to the phrase \textit{``the animal"} in one head, but to the word \textit{``tired"} in the other.

\subsubsection{A Transformer-based language model}

\begin{figure}[h]
\centering
\includegraphics[width=10cm]{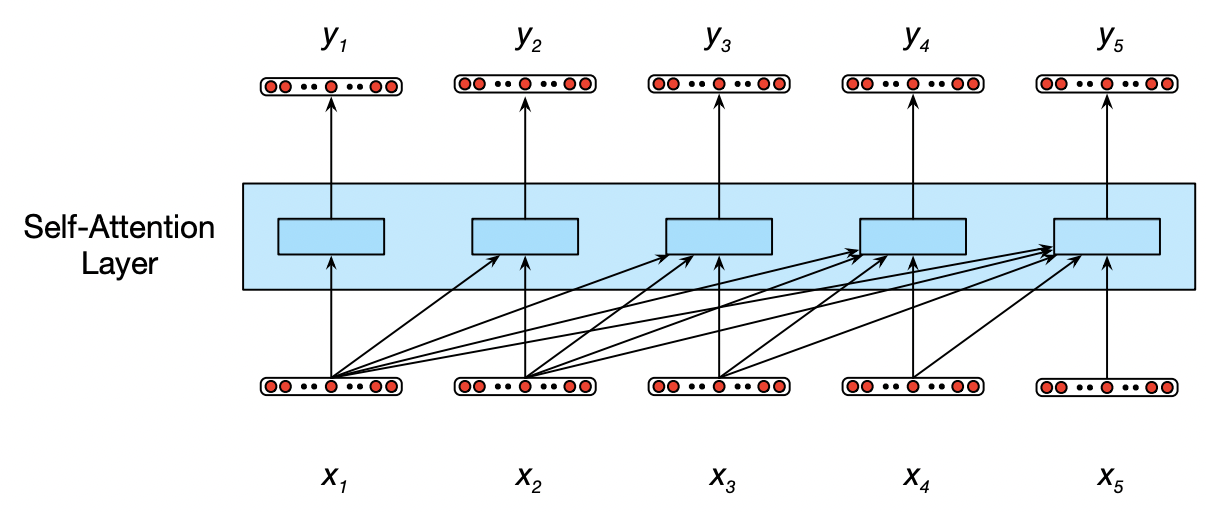}
\caption{Transformer-based language model \citep{jurafsky2009speech}}
\label{TransformerLanguageModel}
\end{figure}

\begin{figure}[t!]
    \centering
    \begin{subfigure}[t]{\textwidth}
        \centering
        \includegraphics[width=12cm]{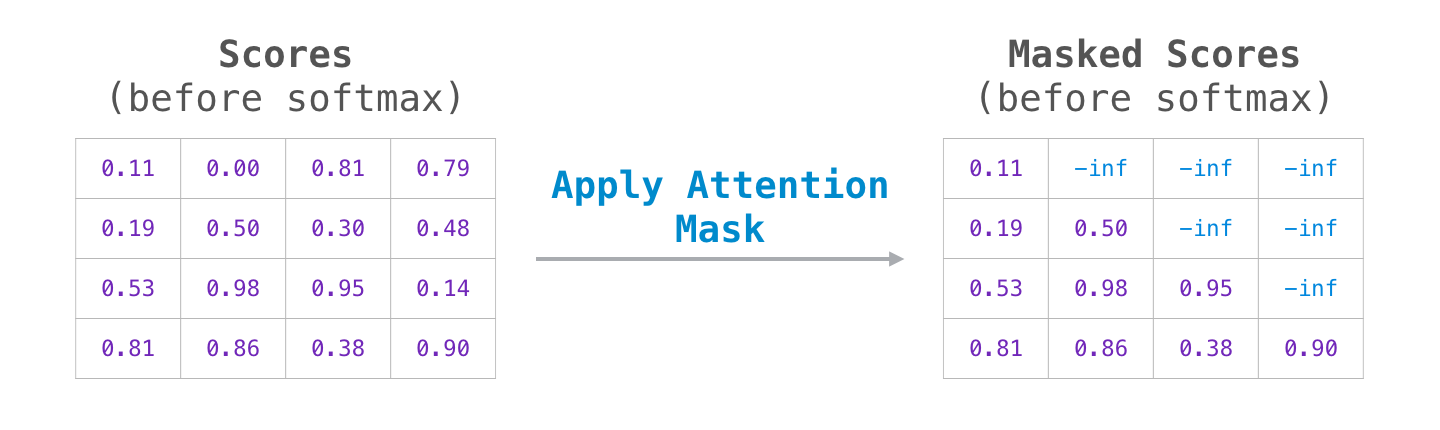}
        \caption{Apply attention mask}
    \end{subfigure}
    \begin{subfigure}[t]{\textwidth}
        \centering
        \includegraphics[width=12cm]{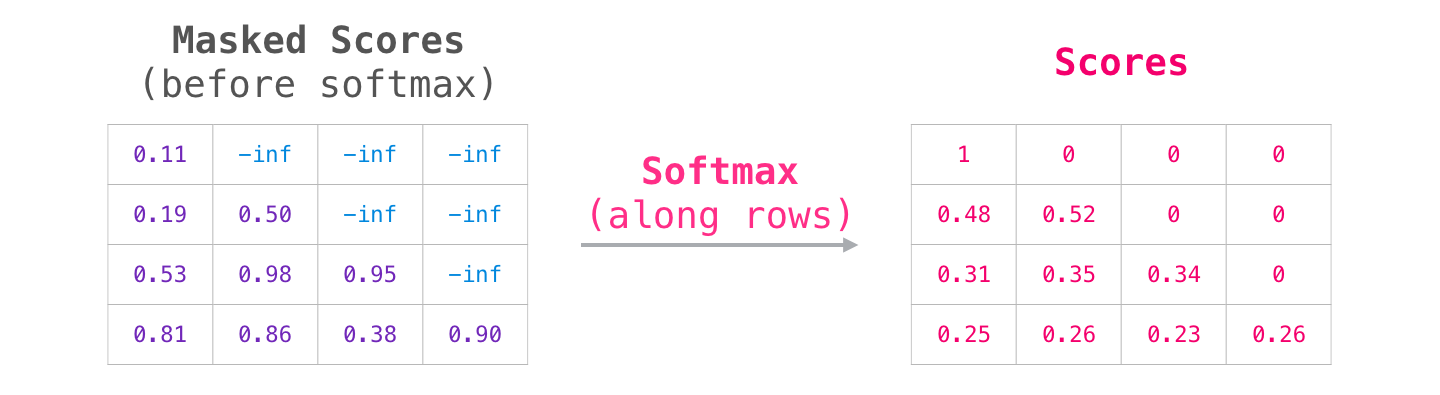}
        \caption{Softmax on masked scores}
    \end{subfigure}
    \caption{Attention mask example. Image credit: \citet{alammar2018transformer}}
    \label{AttentionMaskExample}
\end{figure}

Similar to an RNN language model, a Transformer-based language model can be trained on unlabeled text dataset with next word prediction objective. In every training sequence, each word is passed to the self-attention layer, which allows it to construct its hidden state by attending to all of the previous words (Figure \ref{TransformerLanguageModel}). Different from an RNN network, this computation is independent for all words in the input sequence, meaning that they can be performed in parallel. To prevent a word from attending to its successors, the model makes use of an \textit{attention mask} - a matrix of 0s and 1s to zero out the successors softmax scores (Figure \ref{AttentionMaskExample}). This has enabled Transformer-based language models to be trained on a much bigger scale than RNN and seq2seq models.

\subsection{Pre-trained language models}

\subsubsection{Transformer scalability}

\begin{figure}[h]
\centering
\includegraphics[width=12cm]{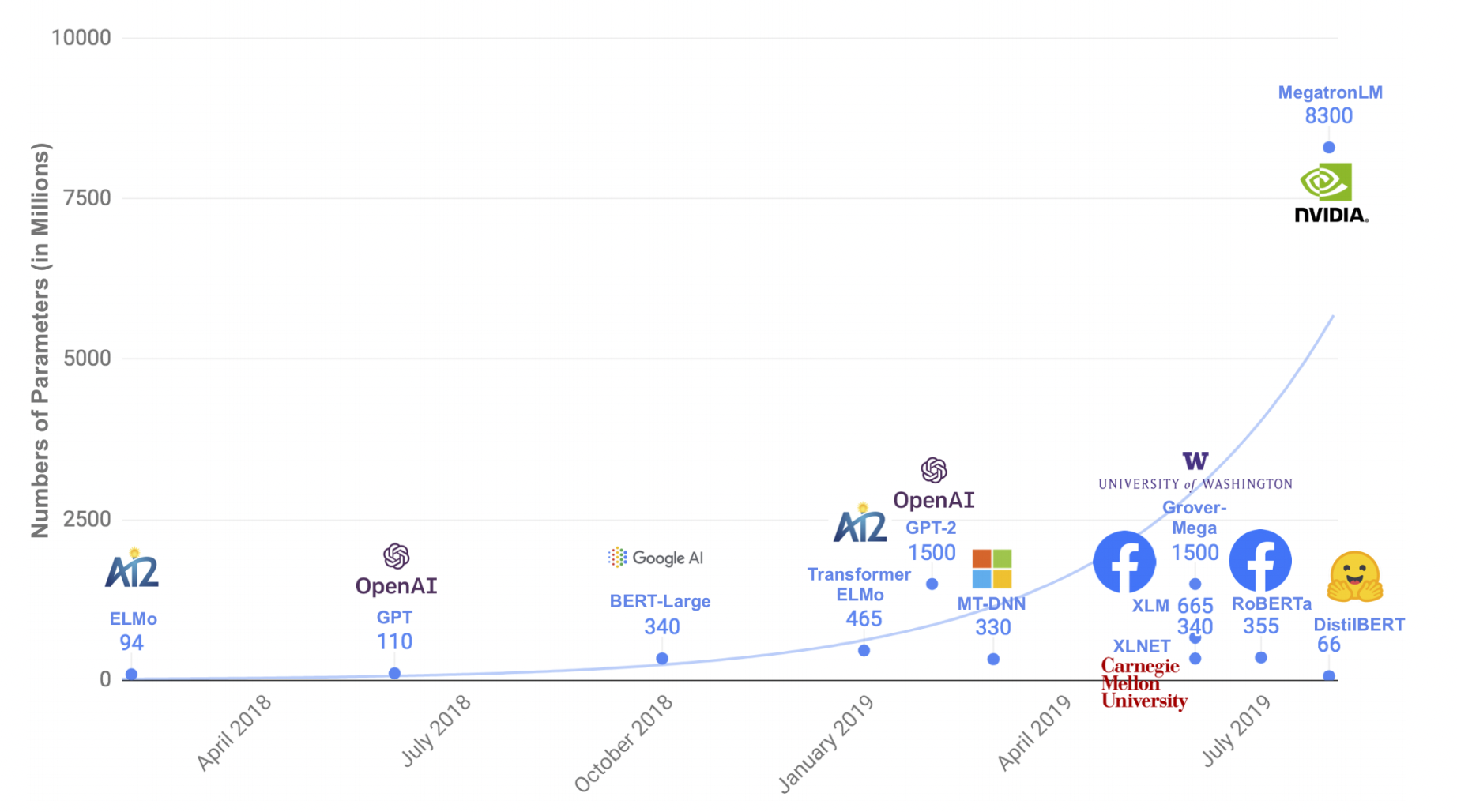}
\caption{Pretrained language models and their sizes \citep{sanh2019distilbert}}
\label{PretrainedLanguageModels}
\end{figure}

Since Transformer models can take advantage of parallel computing resources, language models now can be trained on a massive scale. State-of-the-art pre-trained language models nowadays typically hold up to billions of parameters and are trained on terabytes of unlabeled data \citep{brown2020language}. Figure \ref{PretrainedLanguageModels} shows popular pre-trained language models and their respective sizes. These language models continue to achieve state-of-the-art performance across a plethora of NLP tasks, such as language generation, machine translation or question answering \citep{radford2015unsupervised, radford2019language, peters-etal-2018-deep, yang2019xlnet, devlin-etal-2019-bert, brown2020language}. 

With the advanced performance of Transformer pre-trained language models on various NLP tasks, fine-tuning them has become the go-to approach for doing transfer learning in NLP \citep{howard-ruder-2018-universal, chen-etal-2020-recall, dodge2020fine}. With access to state-of-the-art language models through library like \texttt{Huggingface}\footnote{https://huggingface.co/}, we can simply retrain them on labeled dataset to adapt to specific downstream tasks and achieve competitive results in significantly less time.

The following section gives a brief overview of GPT-2 \citep{radford2019language}, which is what we use to fine-tune all of our experiments in this project.

\subsubsection{GPT-2}

GPT-2 is a Transformer-based language model, with the largest variant containing around 1.5 billion parameters. Similar to the causal language model showed in Figure \ref{TransformerLanguageModel}, GPT-2's objective is to predict the next word given the previous context. GPT-2 was trained on the WebText dataset, which contains 8 million web documents from a variety of domains.

Thanks to huge number of parameters and the diversity of content in its training dataset, GPT-2 is able to achieve impressive performance in an array of language tasks without any supervised training data, including machine translation, summarization, reading comprehension and question answering. Fine-tuning GPT-2 models on specific task has been proved to be an effective approach, as they can achieve competitive results with little training time and supervised signal \citep{ziegler2019fine, see2019massively}.

\section{Neural Text Degeneration}

As discussed, despite of their superiority in multiple NLP tasks, language models are still falling short in open-ended text generation task. With traditional deterministic decoding methods, machine-generated text is often found to be dull, repetitive \citep{holtzman2019curious, welleck2019neural, shao2017generating, fan2018hierarchical}, and sometimes inconsistent and factually incorrect although being fluent and coherent \citep{li2019dont, welleck2019dialogue, hayashi2019latent, petroni2019language} - a problem known as \textit{neural text degeneration}. Example of neural text degeneration with greedy decoding and beam search can be seen in Table \ref{DecodingExample}.

Many studies suspect that the root of this problem might be the maximum likelihood training objective \citep{holtzman2019curious, welleck2019neural}. Multiple alternatives are suggested, which fall into either of the following two categories: using stochastic decoding or tweaking the training objective.

\begin{table}[]
\begin{tabular}{|p{0.2\linewidth} || p{0.8\linewidth}|}
\hline
\textbf{Decoding Method}       & \textbf{Generation} \\ \hline\hline
greedy                         &
The coronavirus is a virus that causes a variety of diseases, including pneumonia, hepatitis, and other respiratory diseases. \textbf{The coronavirus is a virus that causes a variety of diseases, including pneumonia, hepatitis, and other respiratory diseases. The coronavirus is a virus that causes a variety of diseases, including pneumonia, hepatitis, and other respiratory diseases. The coron}\\ \hline\hline

beam search ($b=10$)             & 
COVID-19 is the most common coronavirus in the United States. \textbf{The coronavirus is the most common coronavirus in the United States. The coronavirus is the most common coronavirus in the United States. The coronavirus is the most common coronavirus in the United States.The coronavirus}\\ \hline\hline

temperature sampling ($t=0.7$)   & 
This virus is transmitted from person to person, through the Internet. This infection is characterized by a wide variety of symptoms including fever, headache, fatigue, fatigue, nausea, vomiting, muscle aches, and pain in the chest. The symptoms of COVID-19 include: Decreased heart rate Harmful bowel movements Decreased appetite Increased \\ \hline\hline

top-p ($p=0.9$)                  & 
The outbreak began in October 2008, with 15 cases being confirmed between November 2008 and June 2009. The cases included 11 women aged between 18 and 43 years old. Of these, 1 had died. There is still no evidence of the virus spreading into humans, but it is thought to be spread by people who are infected by infected persons, or infected in public.\\ \hline\hline

top-k ($k=10$)                   & 
The coronavirus is present in about half of all patients and can cause a wide range of health problems including: eye and skin cancers, gastrointestinal problems, and liver and kidney diseases. It is the most widely distributed virus in the United States today, and is transmitted by a number of viruses. It is also the most prevalent and highly contagious virus, causing the death of \\ \hline\hline

penalized sampling ($\theta=1.2$) & 
The virus causes severe respiratory and cardiovascular problems, including death from pneumonia or heart failure in some people with the infection. : A new type of viral that can cause serious health issues such as cancer, hepatitis C, HIV/AIDS, diabetes mellitus, stroke. : An emerging class called ``viruses" are known to be responsible for many diseases like AIDS, malaria, tuberculosis\\ \hline\hline

greedy (with unlikelihood training) &
It is a common and deadly disease that has been linked to the deaths of hundreds of thousands of people.
The first known case of COVID-19 was reported in the late 1960s. The first person to be infected with the virus was a woman named Mary Ann Crouch, who had been living in a house in the village of Crouch.\\ \hline
\end{tabular}
\caption{Example of continuation for the sentence \textit{``Coronavirus disease (COVID-19) is an infectious disease caused by a newly discovered coronavirus."} using GPT-2 Small model with different decoding methods}
\label{DecodingExample}
\end{table}

\subsection{Stochastic Decoding}

All stochastic decoding methods share a common goal, which is to introduce a degree of randomness to the generation process so the model has less chance to repeat itself. In this section, we give an overview of popular stochastic decoding methods in the literature, with their examples of continuations in Table \ref{DecodingExample}.

\subparagraph{Sampling with temperature} 
One can generate more diverse text simply by sampling next words from the learned softmax distribution. This is often done with the use of a Boltzmann temperature parameter \citep{ackley1985learning} to control the randomness of the sampling process, where zero temperature is equivalent to argmax operation, and infinite temperature corresponds to a uniform sampling. However, choosing the right temperature can be tricky, as there is a trade-off between the diversity and quality of the generated text: lower temperatures generate less diverse and predictable words, while higher temperatures generate more diverse text but can produce more implausible words \citep{caccia2018language}.

\subparagraph{Top-k sampling}
\cite{fan2018hierarchical} introduced top-k sampling, which samples from the list of top \textit{k} words with highest probability at each time step, effectively ignoring the tail of the distribution. While top-k sampling has lead to considerably higher quality text compared to sampling with temperature, choosing the right \textit{k} value is still arbitrary: if one word in the list of \textit{k} words make up the most part of the distribution, we are still likely to produce implausible words.

\subparagraph{Nucleus (top-p) sampling} To address the aforementioned problem with top-k sampling, \cite{holtzman2019curious} introduced Nucleus (also known as top-p) sampling, which samples words whose probability mass makes up the top \textit{p} percent of the total distribution. Unlike top-k sampling where only a fixed number of candidates are considered, the number of candidates in Nucleus sampling changes dynamically according to the distribution mass at each time step.

\subparagraph{Penalized sampling} In some context like question answering, sampling the next tokens can lead to a wrong answer since implausible tokens can still receive non-zero probability mass. To take into account this problem, \cite{keskar2019ctrl} proposed penalized sampling, which samples words in a near-greedy fashion but prevents repetitions by discounting the scores of previously generated tokens.

\subsection{Training Objective}

Stochastic decoding methods have one downside: they do not solve the underlying problem with maximum likelihood training. In this section, we examine different strategies to alter language models training objective to cope with neural language degeneration problem.

\subparagraph{Entmax loss/sampling} There exists a mismatch between training and testing conditions in stochastic decoding method, where the model generates text based on a truncated softmax distribution but is evaluated based on the full softmax distribution \citep{martins2020sparse}. The authors thus proposed to use an entmax loss function when training instead of softmax, which transforms a vector of scores into a sparse probability distribution to prevent giving any probability mass to implausible words. At inference time, entmax sampling is used, thus making sure that training and testing conditions are similar. The downside of this approach is that we can no longer use perplexity as an evaluation metric for training language model, since the distribution can contain many zero probability values. To compensate for this drawback, \cite{martins2020sparse} proposed $\epsilon$-perplexity and sparsemax score as alternative evaluation metrics.

\subparagraph{Scheduled sampling}
When training a language model with the maximum likelihood training objective, we have access to the ground-truth tokens at every time step; however, at inference time, we have to rely on tokens generated by the model itself. Thus, the model cannot recover if it makes a single mistake, since it would cascade errors to the whole generated sequence. To address this problem, \cite{bengio2015scheduled} proposed scheduled sampling, in which the model randomly selects between the ground-truth token or model generated token as the label at training time, with the hope that it can learn to correct its own mistakes at inference time. Despite being the winner for the MSCOCO 2015 image captioning challenge, \cite{huszr2015train} suspected that scheduled sampling did not solve the underlying problem of maximum likelihood training and was in fact an inconsistent training strategy. \cite{goyal2017differentiable} revealed that scheduled sampling could not distinguish between local errors and cascading errors as the gradient did not provide enough useful information.

\subparagraph{Generative Adversarial Nets (GAN)}
Since its introduction in \citep{goodfellow2014generative}, GAN has taken over the computer vision community and achieved state-of-the-art results for a plethora of computer vision tasks \citep{radford2015unsupervised, karras2017progressive, brock2018large, zhang2016stackgan}. However, applying GAN to natural language processing is not straightforward since the sampling process cannot be described as a differentiable operation in discrete probabilistic models \citep{huszr2015train}. Several GAN variants have been proposed \citep{yu2017seqgan, guo2017long}; however, till this date, standard MLE models still outperformed GAN models in terms of the quality and diversity of the generated text \citep{caccia2018language}.

\subparagraph{Unlikelihood training}
In addition to the maximum likelihood training objective where the models try to maximize the likelihood of the ground-truth tokens, \cite{welleck2019neural} introduced the unlikelihood training objective, in which the models try to identify negative examples to push down their probability mass. For solving the repetition issue, this is often done in two levels: token level and sequence level. In token level unlikelihood training, at each time step, the model is given a list of negative candidates to calculate the loss, which can be the previous tokens to avoid repetition \citep{welleck2019neural} or a list of tokens that appear too often \citep{li2019dont}. In sequence level unlikelihood training, the model is given a list of prefixes to generate text from, then get penalized for repeated n-grams on its own generation to accommodate for the distribution mismatch between training sequences and generation sequences \citep{welleck2019neural}. For solving the inconsistency issues, \cite{li2019dont} use unlikelihood training on existing natural language inference datasets to penalize contradicting sentence pairs, thus pushing down the probabilities of contradicting utterances.

\section{Evaluation metrics for creative generation}
There are several aspects that we want take a closer look on when evaluating language models on open-ended generation task, which are quality, diversity, and consistency. In this section, we study what metrics have been proposed in the literature and aim to make comparison between them.

\subsection{Quality}

\subparagraph{Corpus-BLEU}
BLEU is originally proposed to evaluate models on machine translation task by comparing the similarity between machine generated translation and human references \citep{papineni2002bleu}. In open-ended text generation task, since we want our models to produce natural and human-like text, BLEU seems like an intuitive metric to use. \citet{yu2017seqgan} proposed the use of BLEU score to judge the quality of machine generated texts by comparing them to a large corpus of human text, which is now being referred in many studies as Corpus-BLEU \citep{caccia2018language, nadeem2020systematic}. To formalize, Corpus-BLEU returns the mean BLEU score of every sample from the set of machine generated text $S_{gen}$ against the whole human reference set $S_{ref}$:

$$\textrm{Corpus-BLEU}(S_{gen}, S_{ref}) = \frac{1}{|S_{gen}|} \sum_{S \in S_{gen}} \textrm{BLEU}(S, S_{ref}) $$

Note that a higher Corpus-BLEU score implies better generation quality since it has more n-gram overlap with the human reference data. The downside of this evaluation metric is its quadratic runtime complexity: for each sample we need to calculate a BLEU score between that sample and the whole reference corpus.

\subparagraph{Forward perplexity}
Because natural, high quality and grammatically correct sentences tend to have higher probabilities than gibberish, we can use the likelihood of a sentence as a proxy for its quality. \citet{zhao2018adversarially} propose the use of a RNN language model that have been trained with real text data to compute the perplexity of a model's samples, which the authors refer to as \textit{forward perplexity}:

$$\textrm{Forward ppl} = \textrm{ppl}_{S_{ref}}(S_{gen}) $$

The reason for using a RNN language model here is to estimate the true distribution of the entire language. This metric can help to measure the fluency of machine generated text which is its quality in essence \citep{zhao2018adversarially, cfka2018eval}. To remove the need for training a RNN language model on human data, one can leverage available pre-trained language models which have been trained on massive dataset such as GPT-2 \citep{radford2019language}.

\subparagraph{Acceptability}
Another way to think about the quality of machine generated texts is their acceptability - how natural they feel to native speakers of the language. Acceptability can be influenced by context: sentences that sound strange when standing alone can appear natural in specific contexts, while those which appear perfectly by themselves may sound odd when surrounding by other sentences \citep{lau2020accept, bizzoni-lappin-2019-effect, bernardy-etal-2018-influence}. This is crucial in open-ended text generation task, since models are usually conditioned on specific contexts before being asked to generate more continuations.

\cite{lau2020accept} propose the use of pre-trained language models to calculate sentence probability as a proxy for its acceptability within a given context. According to the authors finding, using BERT model \citep{devlin-etal-2019-bert} with PenLP \citep{vaswani2017attention} to normalize the sentence's probability produces acceptability scores that match human intuition.

\subsection{Diversity}
\subparagraph{Self-BLEU}
One way to think about diversity is how the generated samples from a collection are different from each other. Using the same intuition as Corpus-BLEU, \citet{zhu2018texygen} introduce Self-BLEU score to access the similarity between every document and the rest of the generated collection. To formalize, Self-BLEU returns the mean BLEU score of every sample from the set of machine generated text $S_{gen}$ against every other samples in the same set $S_{gen}$:

$$\textrm{Self-BLEU}(S_{gen}) = \frac{1}{|S_{gen}|} \sum_{S \in S_{gen}} \textrm{BLEU}(S, S_{gen} \setminus \{S\}) $$

A lower Self-BLEU score implies higher diversity of the collection, as the documents in the collection are more different from each other. Similar to Corpus-BLEU, this metric suffers from its quadratic runtime complexity which makes it intractable for large collection of documents.

\subparagraph{Reverse perplexity/Cross entropy}
Inspired by a similar metric in image generation, \citet{zhao2018adversarially} use an RNN language model to train on generated samples of a model and calculate its perplexity on human data, which the authors refer to as \textit{reverse perplexity}:

$$\textrm{Reverse ppl} = \textrm{ppl}_{S_{gen}}(S_{ref}) $$

Here, the RNN language model resembles the distribution of the generated collection. Similar to how forward perplexity judges the quality of the generated collection using human data, reverse perplexity measures the quality of human data based on the generated collection. If the generated collection is diverse enough to represent different writing styles or topics, the RNN language model should perceive human text as \textit{natural} and \textit{fluent}, therefore giving low perplexity to human data.

\subparagraph{Sequence repetition}
While human rarely repeat themselves when writing, machine generated texts are often found to be repetitive, especially when being produced by deterministic decoding methods \citep{holtzman2019curious, welleck2019neural, shao2017generating, fan2018hierarchical}. While not being diverse does not necessarily mean being repetitive, being repetitive prevents the model from generating diverse continuations. Therefore, using repetition as a metric can give us an idea how diverse a document collection is. \citet{welleck2019neural} use a metric called \textit{seq-rep-n} to measure sequence repetition by calculating the portion of duplicate n-grams in a generated sequence $S$:

$$\textrm{seq-rep-n} = 1.0 - \frac{|\textrm{unique n-grams}(S)|}{|\textrm{n-grams}|} $$

\subsection{Consistency/Commonsense reasoning}
\label{Consistency/Commonsense reasoning Lit Review}
One of the biggest challenge for a model in open-ended text generation tasks is to demonstrate its commonsense/logical reasoning ability. This is similar to the task of natural language inference (NLI), in which the model is given pairs of sentences and must decide whether the relationship between them are neutral, entailment or contradiction \citep{fyodorov2000natural, condoravdi2003entailment, bos2005recognising, dagan2005pascal, maccartney2009extended}. 

To evaluate their dialogue generation model's commonsense reasoning ability, \citet{li2019dont} calculates its \textit{selection accuracy} against the Dialogue NLI dataset \citep{welleck-etal-2019-dialogue}, in which they assume the model would select the sentence with lower perplexity. By using a similar approach, we can evaluate any generative model in presence of an NLI dataset without the need to turn it into a classification model. Depending on the ended-goal task, there exists a number of NLI dataset where this metric can be used:

\subparagraph{MultiNLI}
 \citet{williams2018nli} introduce the Multi-Genre NLI Corpus (MultiNLI), which is made up from a variety of writing and spoken materials, ranging from reports, speeches, letters, conversations to non-fiction and fiction books. Due to its diversity in writing styles, genres and topics, this dataset is suitable for many open-ended text generation tasks.

\subparagraph{Dialogue NLI}
\cite{welleck-etal-2019-dialogue} introduce the Dialogue NLI dataset, which contains labeled pairs of sentences constructing from the Persona-Chat dialogue dataset \citep{zhang2018personalizing}. This dataset is most suitable for dialogue generation task.

\subparagraph{ROCStories}
\citet{mostafazadeh2016corpus} propose a evaluation framework called the Story Cloze Test, where the model has to select the correct ending to a four-sentence story. The corpus - ROCStories - is divided into two parts: a training set containing full five-sentence stories, and the test set consisting of four opening sentences of the stories with two different endings. The stories are designed so that the they contain a mixture of commonsense causal and temporal relations between daily events. This dataset is most suitable for story generation task.

\lhead{\emph{How to Evaluate Creative Generation?}}
\chapter{How to Evaluate Creative Generation?}

It is clear that using only traditional metrics such as perplexity is not enough to evaluate language models on open-ended text generation task. Instead, we need to look at their performance on all three dimensions: quality, diversity and consistency of the generated text.

In this chapter, we first present our experiment to compare different evaluation metrics for each of the dimension when evaluating language models on open-ended text generation task. We then decide on what is the best metric to use for each dimension, and use those metrics to assess the two common techniques for solving neural text degeneration: stochastic decoding methods and tweaking the training objective. In this thesis, we explore using unlikelihood training \citep{welleck2019neural} as a representative for the training objective strategy.

\section{Setup}

\subsection{Dataset}
We focus on story generation, an instance of open-ended generation, in this thesis. For this reason, we decide to use the Harry Potter series by J.K. Rowling as our training corpus. We use a version of the whole series from \textit{https://github.com/joycex99/hp-word-model}, which has been striped off of page numbers and headings.

The corpus contains 1,104,770 words in total (1,710,720 subwords), with 29,245 unique words (50,257 unique subwords). We use the GPT-2 Tokenizer\footnote{https://huggingface.co/transformers/model\_doc/gpt2.html\#gpt2tokenizer} to tokenize the text into sequences of 200 subwords in length. We do a train/dev/test split ratio of 80/10/10, which results in 6,220 sequences for training, 778 sequences for development and 778 sequences for testing.

\subsection{Unlikelihood training}

The main goal of unlikelihood training is to push down the probability mass of \textit{negative candidates}. Given a sequence of tokens $(x_1,...,x_T)$ and a set of negative candidates at time step $t$ $C^t = \{c_1,...,c_m\}$, \citet{welleck2019neural} define the unlikelihood loss at time step $t$ as:

$$
\mathcal{L}_{UL}(p_{\theta}(\cdot|x_{<t}), C^t) = - \sum_{c \in C^t} \textrm{log}(1 - p_{\theta}(c|x_{<t}))
$$

The unlikelihood loss can be used along side the usual cross-entropy loss when training/fine-tuning the language model. The unlikelihood loss can be applied at two levels: token-level and sequence-level.

\paragraph{Token-level loss} For token-level loss, the list of negative candidates at each time step is all of the tokens from previous time steps, so the model can avoid repeating tokens that it has seen before \citep{welleck2019neural}.

\paragraph{Sequence-level loss} The token-level loss is restricted to negative candidates selected from the training distribution. To accommodate for the difference between training and decoding condition, \citet{welleck2019neural} proposed the use of a sequence-level loss, where we ask the model to generate continuations given some contexts, then take the repeated n-grams from the continuations as negative candidates.

\subsection{Training}

We use the pre-trained GPT-2 Small from \texttt{HuggingFace}\footnote{https://huggingface.co/transformers/model\_doc/gpt2.html\#gpt2lmheadmodel} as our base model. We fine-tune it with the Harry Potter books dataset using two different training methods: maximum likelihood estimate (MLE) training and unlikelihood (UL) training \citep{welleck2019neural}. With unlikelihood training, we follow what the authors have suggested: with probability of 0.5 use sequence-level loss otherwise use the token-level loss. To compute sequence-level loss (with n-gram of 4), we use the prefix of size 50 of each training sequence in the current training batch and greedily decode continuations of length 100. All models are trained with 4 epochs using Adam optimizer with batch size of 12 and learning rate of 0.001. To find out how sensitive unlikelihood training is to the number of training epochs, we also train another model using unlikelihood training with 1 epoch only.

We are also interested in which effects do larger models have on text generation. Whenever possible, we repeat the experiments with a GPT-2 Medium base model using a similar training regime.

\section{Results}

\subsection{Similarity to human text}

In this section, we want to investigate how different the token distributions of human text are from that of machine-generated text. Our hypothesis is that if one model can produce text which have a similar distribution of tokens to human text, it would read natural and human-like. Below is a paragraph that we have taken from the Harry Potter series:

\begin{quote}
    Harry, who was shaking all over, thought for a moment that Dumbledore might not be able to climb into the boat; he staggered a little as he attempted it; all his efforts seemed to be going into maintaining the ring of protective flame around them. Harry seized him and helped him back to his seat. Once they were both safely jammed inside again, the boat began to move back across the black water, away from the rock, still encircled by that ring of fire, and it seemed that the Inferi swarming below them did not dare resurface.

    ``Sir," panted Harry, ``sir, I forgot -- about fire -- they were coming at me and I panicked --"
    
    ``Quite understandable," murmured Dumbledore. Harry was alarmed to hear how faint his voice was.
    
    They reached the bank with a little bump and Harry leapt out, then turned quickly to help Dumbledore. The moment that Dumbledore reached the bank he let his wand hand fall
\end{quote}

\begin{figure}[t!]
    \centering
    \begin{subfigure}[t]{0.5\textwidth}
        \centering
        \includegraphics[width=7cm]{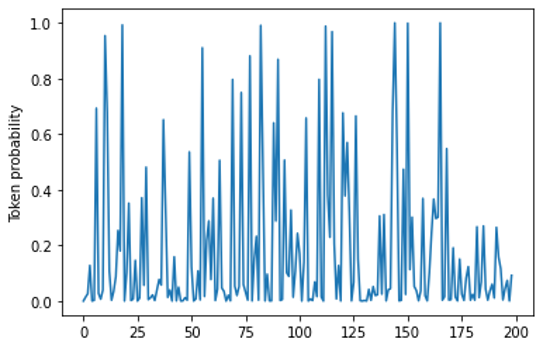}
        \caption{GPT-2}
        \label{GPT-2 human prob}
    \end{subfigure}%
    ~
    \begin{subfigure}[t]{0.5\textwidth}
        \centering
        \includegraphics[width=7cm]{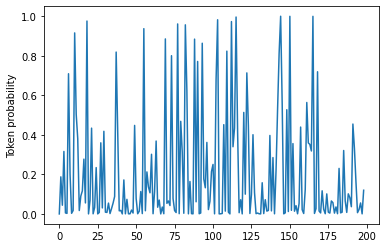}
        \caption{MLE GPT-2}
        \label{MLE GPT-2 human prob}
    \end{subfigure}
    ~
    \begin{subfigure}[t]{0.5\textwidth}
        \centering
        \includegraphics[width=7cm]{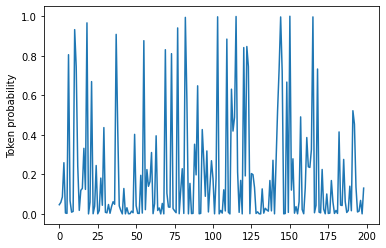}
        \caption{UL GPT-2 (1 epoch)}
        \label{UL GPT-2 (1 epoch) human prob}
    \end{subfigure}%
    ~
    \begin{subfigure}[t]{0.5\textwidth}
        \centering
        \includegraphics[width=7cm]{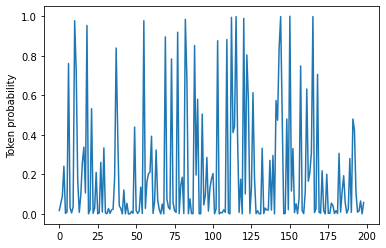}
        \caption{UL GPT-2 (4 epochs)}
        \label{UL GPT-2 (4 epochs) human prob}
    \end{subfigure}%
    \caption{Distribution of tokens for the human written paragraph}
    \label{HumanTextDistribution}
\end{figure}

 We then use the GPT-2 Small model as an oracle to calculate the probability of each token in the paragraph (Figure \ref{HumanTextDistribution}). From the plot, we can see that the distribution of human written text does not follow a consistent pattern but moves up and down arbitrarily, which means when writing, human never try to find the most probable word after every word. This explains why deterministic decoding methods such as greedy and beam search fail to produce natural and human-like text.

We use the same prefix with the above paragraph (\textit{Harry, who was shaking all over}) and generate continuations of similar length (200 subwords) using our three models: Maximum Likelihood Estimate GPT-2 (MLE GPT-2), Unlikelihood Training with 4 epochs (UL GPT-2 4 epochs), and Unlikelihood Training with 1 epoch (UL GPT-2 1 epoch). All continuations are generated using greedy decoding, except for the MLE model where we also experiment with top-p and top-k decoding, with the value p set to 0.9 and the value k set to 10. The machine generated continuations are given below:

\begin{quote}
    \textbf{MLE GPT-2, greedy decoding}: Harry, who was shaking all over, was still standing there, looking at the ceiling.
    
    ``I'm sorry," he said, ``but I'm not going to tell you what happened. I'm not going to tell you what happened. I'm not going to tell you what happened. I'm not going to tell you what happened. I'm not going to tell you what happened. I'm not going to tell you what happened. I'm not going to tell you what happened. I'm not going to tell you what happened. I'm not going to tell you what happened. I'm not going to tell you what happened. I'm not going to tell you what happened. I'm not going to tell you what happened. I'm not going to tell you what happened. I'm not going to tell you what happened. I'm not going to tell you what happened. I'm not going to tell you what happened. I'm not going to tell you what happened.
\end{quote}

\begin{quote}
    \textbf{MLE GPT-2, top-p decoding, p = 0.9}: Harry, who was shaking all over and watching, said, ``You don't understand how to make a mistake --"

    ``I am not a wizard," said Harry. ``I do not believe that my mother's magical work was a good idea.
    
    But she was, after all, a wizard...."
    
    ``That's right, I doubt she was," said Hermione, pulling a few strands of parchment out of her sleeve and peering at them.
    
    ``And that's a sign, then," said Harry. ``What are you, then?"
    
    ``You," said Hermione, with a look of great horror. ``I have a badge for my mother..."
    
    ``I have a badge for my mother." Harry, who was standing behind her, was now smiling with his forehead on his robes.
    
    ``What does it say?"
    
    ``It tells you..."
    
    The door of the tent opened and
\end{quote}

\begin{quote}    
    \textbf{MLE GPT-2, top-k decoding, k = 10}: Harry, who was shaking all over himself, and Harry looked up to see his father, Ron, who was still looking at him from his seat.

    ``Well, you'll see, I mean, it's been a long time," Ron said, looking down at his hands. ``But it's not like I'm telling you anything, isn' it, isn' it? You can't say you were a bit scared, or you're just not doing the right thing. You can't."
    
    Harry looked up again, but Ron had been right. The look in their eyes was as much relief as it had been feeling. Harry thought he saw a bit of remorse in the look in Ron's face.
    
    ``You're sorry," said Hagrid, ``but I've just got a feeling something will happen, Harry Potter. It's not like we've got all these other people's names on it — you don't know anything about the Ministry of Magic."
\end{quote}

\begin{quote}
    \textbf{UL GPT-2 (1 epoch), greedy decoding}: Harry, who was shaking all over the place, was a little bit worried that he might be able to get away with it.

    ``You're not going to be able to do that," said Harry, looking at the door, which was open to the outside. Harry had to stop himself, because he was still looking at the door with a very large smile on his face. He had a feeling that he was going to be able to get away with it, though, because he had a very large hand on the wand, and Harry was not sure whether he was going to be able to see the door open again until the very end of the night.
    
    Harry had to go back to the bathroom, and when he was done, he was looking at the door again, and Harry felt very lucky. He had been able to see the door open for a moment, and then he saw the door open again, and he was very glad that he had managed to get inside.

\end{quote}
 
\begin{quote}
    \textbf{UL GPT-2 (4 epochs), greedy decoding}: Harry, who was shaking all over the place.

    ``I'm sorry," said Harry, ``but I'm not going to tell you what to do next time. I'm just going to tell you what I want to see Dumbledore do when he's here."
    -Ron Weasley, who was now looking back at Dumbledore's portrait, and the Death Spell wandcrossing into the Forbidden Forest Firep. Translator's Transmogrified Spellbag -``And now, Harry, you can tell the whole world about the Patronuses Curse Curse Patron-Blood Lord's friends, Ron, Hermione, and Sirius face on Deathly Deathly High Inquisitor's face-Ex-Ex-Lionbeer. Patron Patron:  Dumbledore Death Spellpike-GentorHistoryLionbus AbridgedAndHarry's parents, Ron, Galleons, and Neville.
\end{quote}

\begin{figure}[t!]
    \centering
    \begin{subfigure}[t]{0.5\textwidth}
        \centering
        \includegraphics[width=7cm]{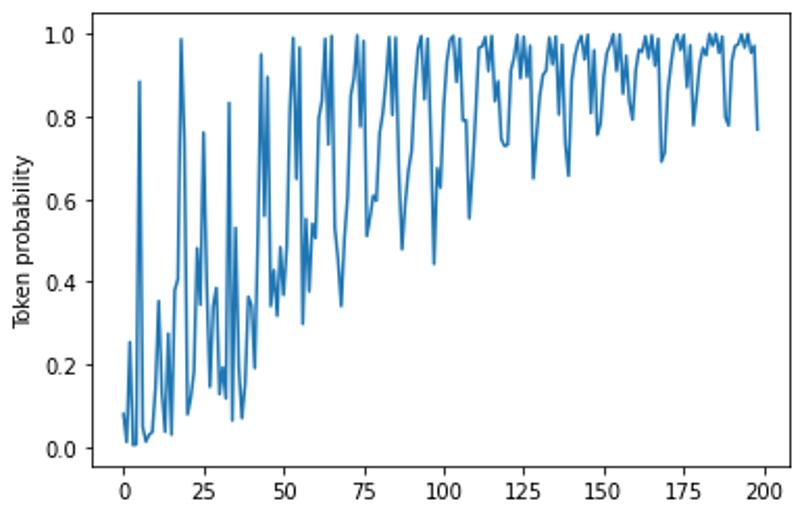}
        \caption{MLE GPT-2, greedy decoding}
        \label{MLE GPT-2, greedy decoding}
    \end{subfigure}%
    ~
    \begin{subfigure}[t]{0.5\textwidth}
        \centering
        \includegraphics[width=7cm]{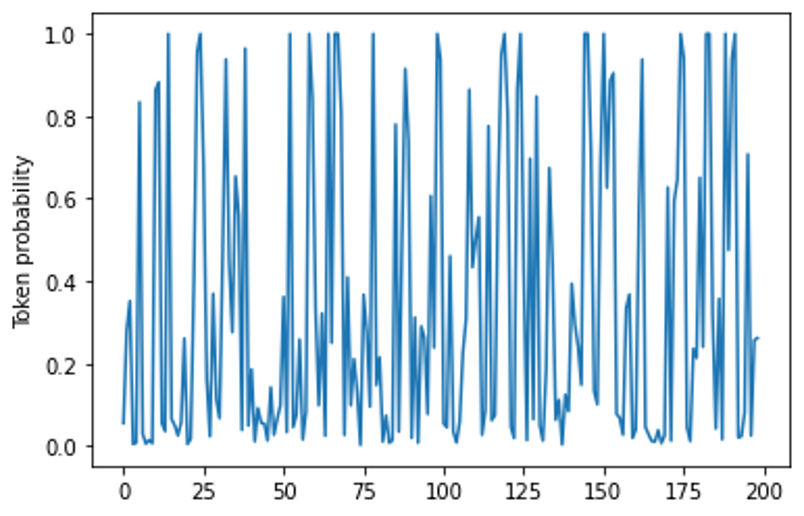}
        \caption{MLE GPT-2, top-p decoding, p = 0.9}
        \label{MLE GPT-2, top-p decoding, p = 0.9}
    \end{subfigure}%
    ~
    \vskip\baselineskip
    \begin{subfigure}[t]{0.5\textwidth}
        \centering
        \includegraphics[width=7cm]{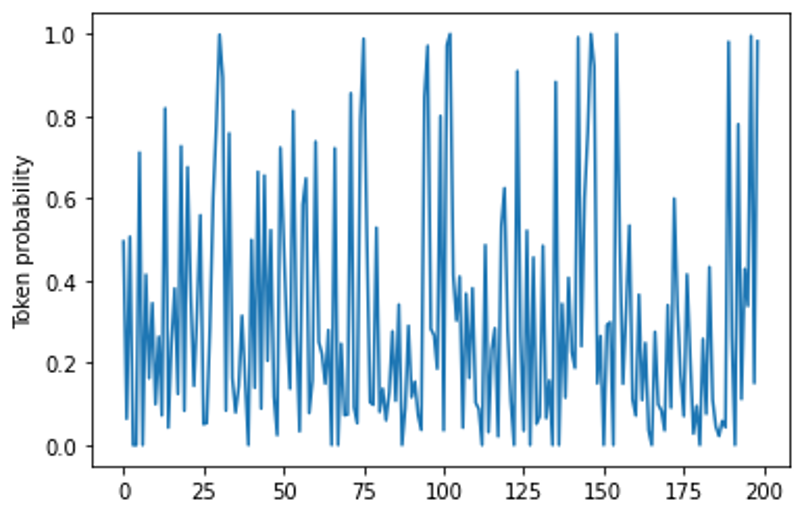}
        \caption{MLE GPT-2, top-k decoding, k = 10}
        \label{MLE GPT-2, top-k decoding, k = 10}
    \end{subfigure}
    ~
    \begin{subfigure}[t]{0.5\textwidth}
        \centering
        \includegraphics[width=7cm]{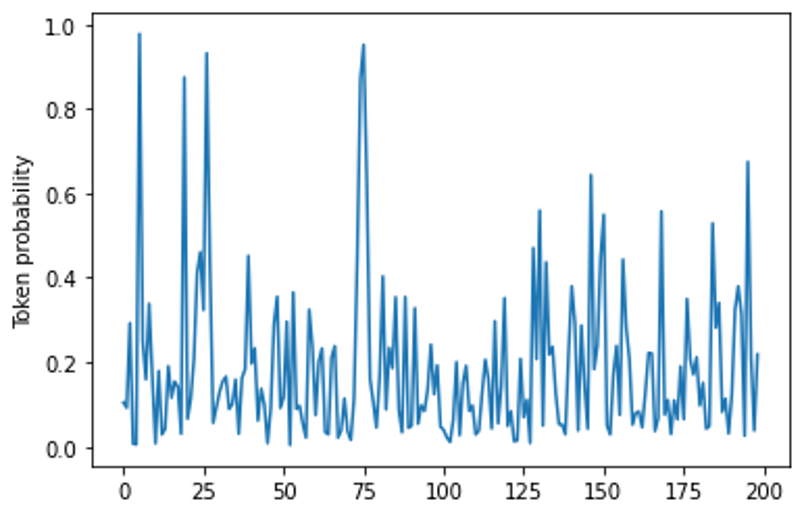}
        \caption{UL GPT-2 (1 epoch), greedy decoding}
        \label{UL GPT-2 (1 epoch), greedy decoding}
    \end{subfigure}%
    ~
    \begin{subfigure}[t]{0.5\textwidth}
        \centering
        \includegraphics[width=7cm]{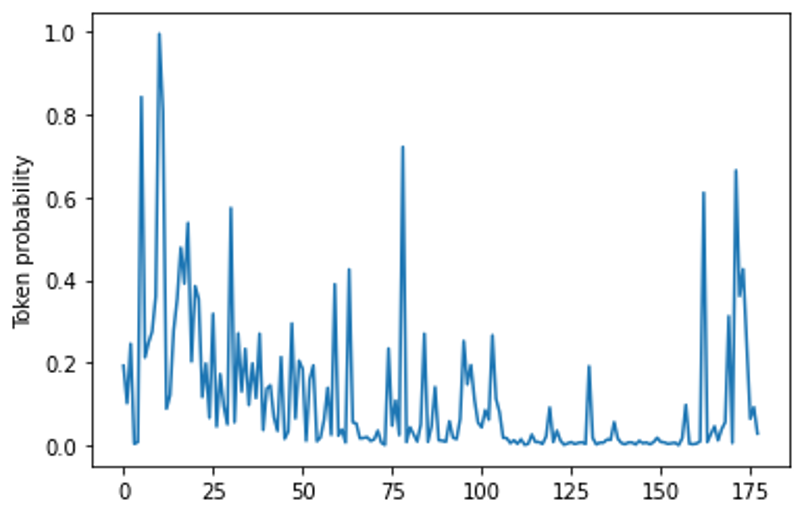}
        \caption{UL GPT-2 (4 epochs), greedy decoding}
        \label{UL GPT-2 (4 epochs), greedy decoding}
    \end{subfigure}%
    \caption{Distribution of tokens for the machine generated paragraphs}
    \label{MachineTextDistribution}
\end{figure}

We then calculate the token distribution of each paragraph using the same model which generates it. For continuations that are generated using stochastic decoding, we augment the distribution to match with the behavior of that decoding method: at each time step we only consider the top 90\% of the tokens for top-p decoding and top 10 tokens for top-k decoding, then adjust the probability accordingly so the total probabilities sum up to 1. The plot of the distribution is given in Figure \ref{MachineTextDistribution}.

The MLE GPT-2 with greedy decoding (Figure \ref{MLE GPT-2, greedy decoding}) has observed a phenomenon called \textit{model overconfidence} \citep{see2019massively, jiang-de-rijke-2018-sequence, holtzman2019curious}, where the model is getting more confident over time and place high probability on a small number of tokens. This explains why it keeps generating the sentence \textit{I'm not going to tell you what happened} again and again.

Out of the five paragraphs, the two generated by MLE GPT-2 with top-p (Figure \ref{MLE GPT-2, top-p decoding, p = 0.9}) and top-k (Figure \ref{MLE GPT-2, top-k decoding, k = 10}) decoding have the most similar distribution to the given human text. Although in terms of commonsense and logical reasoning, the paragraphs do not really make sense, but still they feel more native to human written language than MLE GPT-2 with greedy decoding.

Compare to the other samples, the two paragraphs generated by Unlikelihood training models have quite low probabilities overall (Figure \ref{UL GPT-2 (1 epoch), greedy decoding} and \ref{UL GPT-2 (4 epochs), greedy decoding}), especially the one being trained with 4 epochs where most tokens have a probability value smaller than 0.2. With the UL GPT-2 (4 epochs) model, the generated paragraph stops making sense after the first one hundred subwords and starts to produce only gibberish. The one being generated by UL GPT-2 (1 epoch) model is great in terms of quality, however it is still repetitive in a sense, as the story keeps repeating some of the phrases like \textit{able to do that}, \textit{able to get away with it} and \textit{see the door open}. This suggests unlikelihood training can be difficult to train as it is quite sensitive to the number of training epochs.

\subsection{Quality vs. Diversity}

In this section, our main objectives are (i) finding the best metrics to evaluate quality and diversity of machine generated text and (ii) comparing the models based on their  quality/diversity trade-off.

Since there are a number of parameters to tweak, .i.e the choice of decoding method or the choice of parameters in stochastic decoding method, it can be hard to determine which \textit{model} are superior in text generation. \citet{caccia2018language} proposes a method called \textit{temperature sweep}, which use a fixed set of parameters across all models then see which model has the best result \textit{overall}. Following a similar approach, for each model we generate continuations using three decoding algorithms: greedy, top-p and top-k decoding. For the hyper-parameter p and k, we select a fixed set of values for p from the range of $[0.2...0.96]$ and a fixed set values for k from the range of $[2...400]$. To generate the model samples, we take the prefix of length 50 subwords of every sequence from the training set and generate continuations using different decoding methods/parameters of length 150 subwords. In total, we have 57 pairs of model/parameter, each generating 6,220 different samples.

We use all of the evaluation metrics as discussed in Chapter 2. For Corpus-BLEU and Self-BLEU, we only select 500 continuations from each sample set to calculate the scores due to their runtime complexity. For reverse perplexity, instead of an RNN language model as the authors have suggested, we use a GPT-2 Small model to speed up the training process. We fine-tuned it on the 6,220 generated samples of each model/parameter pair, then calculate the perplexity against our Harry Potter corpus test set. For forward perplexity, we use an off the shelve GPT-2 Small model to calculate the perplexity against the generated sample set of each model/parameter pair.

\subsubsection{Comparison of evaluation metrics}

\begin{figure}[t!]
    \centering
    \begin{subfigure}[t]{0.5\textwidth}
        \centering
        \includegraphics[width=7cm]{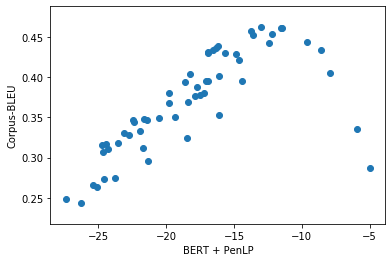}
        \caption{BERT + PenLP vs. Corpus-BLEU}
        \label{BERT + PenLP vs. Corpus-BLEU}
    \end{subfigure}%
    ~
    \begin{subfigure}[t]{0.5\textwidth}
        \centering
        \includegraphics[width=7cm]{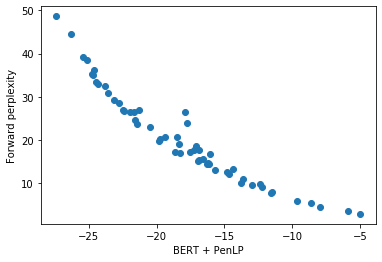}
        \caption{BERT + PenLP vs. Forward perplexity}
        \label{BERT + PenLP vs. Forward perplexity}
    \end{subfigure}

    \begin{subfigure}[t]{0.5\textwidth}
        \centering
        \includegraphics[width=7cm]{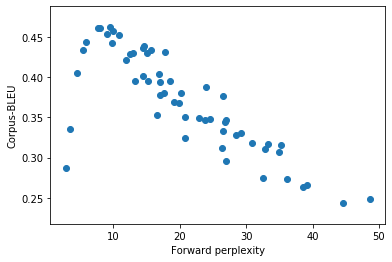}
        \caption{Corpus-BLEU vs. Forward perplexity}
        \label{Corpus-BLEU vs. Forward perplexity}
    \end{subfigure}%
    \caption{Scatter plots between Corpus-BLEU, Forward perplexity and BERT + PenLP}
    \label{QualityMetricComparison}
\end{figure}

Figure \ref{QualityMetricComparison} shows the relationship between the three evaluation metrics for judging quality. BERT + PenLP and Forward perplexity highly correlates with each other overall (Figure \ref{BERT + PenLP vs. Forward perplexity}), and they both agree with Corpus-BLEU up until an inflection point where we observe a reverse trend (Figure \ref{BERT + PenLP vs. Corpus-BLEU} and \ref{Corpus-BLEU vs. Forward perplexity}). The points where the reverse trend happens are either generated by (i) greedy decoding method or (ii) stochastic decoding method with low values for p and k which behaves in a near greedy manner. This is similar to \textit{the likelihood trap} phenomenon \citep{zhang2020trading}, in which the authors observe that human judgement is positively related with the log likelihood of the model up to a certain point where it becomes negatively related afterward. Based on this finding, we argue that Corpus-BLEU is the best metric to evaluate the quality of generated text, since it behaves more like \textit{human judgement} and is able to detect mode collapse.

\begin{figure}[t!]
    \centering
    \begin{subfigure}[t]{0.5\textwidth}
        \centering
        \includegraphics[width=7cm]{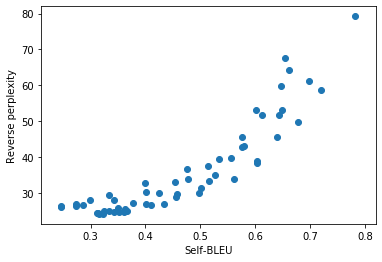}
        \caption{Self-BLEU vs. Reverse perplexity}
    \end{subfigure}%
    ~
    \begin{subfigure}[t]{0.5\textwidth}
        \centering
        \includegraphics[width=7cm]{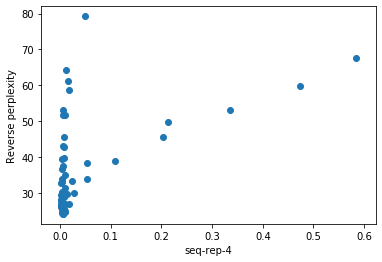}
        \caption{seq-rep-4 vs. Reverse perplexity}
    \end{subfigure}

    \begin{subfigure}[t]{0.5\textwidth}
        \centering
        \includegraphics[width=7cm]{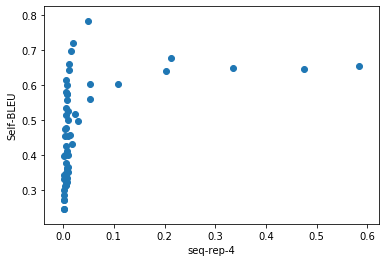}
        \caption{seq-rep-4 vs. Self-BLEU}
    \end{subfigure}%
    \caption{Scatter plots between Self-BLEU, Reverse perplexity and seq-rep-4}
    \label{DiversityMetricComparison}
\end{figure}

Figure \ref{DiversityMetricComparison} shows the relationship between the three evaluation metrics for evaluating diversity. Self-BLEU and Reverse perplexity generally correlates well with each other. seq-rep-4 is able to detect highly repetitive samples; however, it is harder to see how difference the models performance are, since the magnitudes are marginal. As for complexity, using reverse perplexity can be much slower compared to the other metrics as we have to train a language model with the generated samples. Therefore, we choose to use Self-BLEU for the rest of the experiments.

\subsubsection{Quality/diversity trade-off}

\begin{figure}[h]
\centering
\includegraphics[width=14cm]{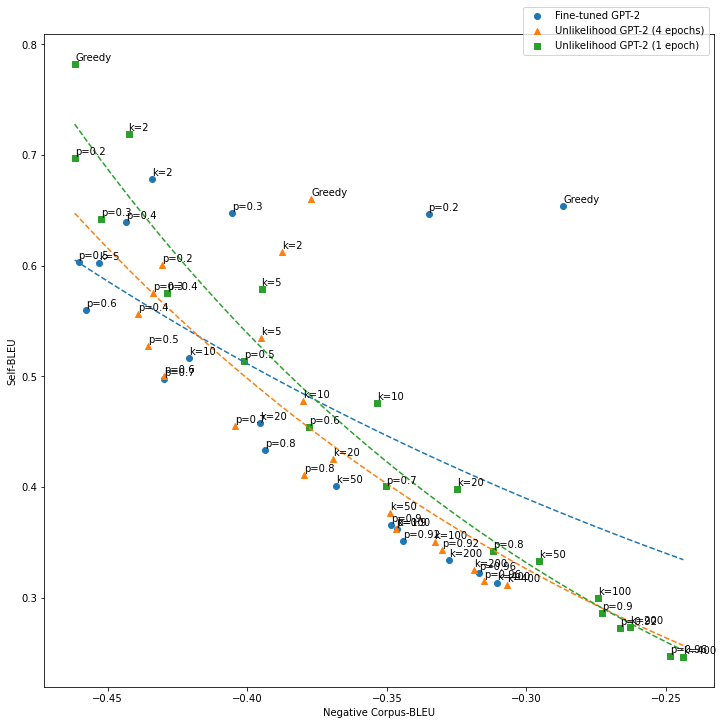}
\caption{Quality vs. Diversity trade-off between models}
\label{QualityDiversityTradeOff}
\end{figure}

Using Corpus-BLEU and Self-BLEU as the two evaluation metrics for quality and diversity, we plot the quality/diversity trade-off between the models in Figure \ref{QualityDiversityTradeOff} and use a log function to fit the data points. Note that we use Negative Corpus-BLEU instead of Corpus-BLEU, so that the lower is better for both metrics. Looking at the graph, it is unclear which model is the best in the quality-diversity trade-off space, since they all lie similar diagonal lines. The trade-off between quality and diversity is clear: increasing the degree of randomness (by decreasing p in top-p decoding or increasing k in top-k decoding) improves diversity but worsens quality and vice versa.

Even though the samples generated by MLE GPT-2 with greedy decoding are much more repetitive (with seq-rep-4 of 0.584 and Self-BLEU of 0.65), UL GPT-2 (1 epoch) with greedy decoding appears much worse in terms of diversity with a Self-BLEU score of 0.78 and seq-rep-4 of 0.004 (Table \ref{SeqRep4Result} and Figure \ref{QualityDiversityTradeOff}). Taking a closer look at the generated samples, we found out that no matter what context were given to the model, UL GPT-2 still produces texts that are rather similar. Below are several examples:

\begin{quote}
``But the task's not till tonight!"  said Harry, accidentally spilling scrambled eggs down his front, afraid he had mistaken the time.

``I'm aware of that, Potter," she said.

\textbf{``But I'm not going to tell anyone. I've got to go and find the Ministry of Magic, and I'm not going to be in the Ministry's office for the rest of the year. I've got a lot more to do with the Ministry's Head of Magical Law Enforcement than I could have ever expected."}

\textbf{Harry looked around the room, and then back to the Ministry's office. He had not seen the Ministry's office for a long time, but he had to admit}
\end{quote}

\begin{quote}
``Well, isn't it obvious?" said Hermione, with a look of maddening superiority.

``If you don't want to tell us, don't," snapped Ron.

``Fine," said Hermione haughtily. 

\textbf{``I'll just tell you what, then. I'll be back in a few minutes."}

\textbf{The train was a long way from the station, and Harry was still not sure whether he was going to be able to get back to the castle. He had to wait until the next morning, when he had a chance to ask the old man if he wanted to go back to Hogwarts.}

\textbf{``I'll see you at the Ministry," he said, and he walked back to the}
\end{quote}

\begin{quote}
``COME BACK, YOU THIEVING --!"

``There's no point, Harry." Tonks had appeared out of nowhere, her mousy hair wet with \textbf{sweat. ``You're going to have to get back to Hogwarts, you know that, if you're going to be there, you'll have to be in the Ministry's office. You've got to get your wand back, and I'll be there, and I'll be there for you, Harry Potter, and you'll have a very special place in the school."}

\textbf{Harry looked around the room, his eyes wide open, and saw the Professor's face. He was wearing a very}
\end{quote}

We can see that even though the contexts are different, the continuations always drift to \textit{Ministry of Magics} and \textit{getting back to Hogwarts}. This proves that looking at repetition as a metric alone is not enough to judge the diversity of generated text, and although unlikelihood training might solve the problem of generating repetitive text, it does not help the model to produce more diverse text.

\begin{table}[]
\centering
\begin{tabular}{|c||c|c|c|}
\hline
                & \textbf{MLE GPT-2} & \textbf{UL GPT-2 (1 epoch)} & \textbf{UL GPT-2 (4 epochs)} \\ \hline\hline
\textit{Greedy} & \textbf{0.584}     & 0.004                       & 0.011                        \\ \hline
\textit{p=0.2}  & \textbf{0.474}     & 0.016                       & 0.007                        \\ \hline
\textit{p=0.4}  & \textbf{0.203}     & 0.008                       & 0.008                        \\ \hline
\textit{p=0.6}  & 0.054              & 0.004                       & 0.010                        \\ \hline
\textit{p=0.8}  & 0.017              & 0.003                       & 0.008                        \\ \hline
\textit{p=0.9}  & 0.010              & 0.002                       & 0.007                        \\ \hline
\textit{k=2}    & \textbf{0.213}     & 0.019                       & 0.006                        \\ \hline
\textit{k=10}   & 0.024              & 0.004                       & 0.005                        \\ \hline
\textit{k=50}   & 0.010              & 0.002                       & 0.005                        \\ \hline
\textit{k=100}  & 0.009              & 0.002                       & 0.005                        \\ \hline
\textit{k=200}  & 0.007              & 0.002                       & 0.005                        \\ \hline\hline
\textit{Human Text*} & \multicolumn{3}{c|}{0.007}                        \\ \hline
\end{tabular}
\caption{Sequence repetition scores between different models and decoding methods. *Human text sequence repetition is computed using the training dataset}
\label{SeqRep4Result}
\end{table}

\begin{figure}[h]
\centering
\includegraphics[width=10cm]{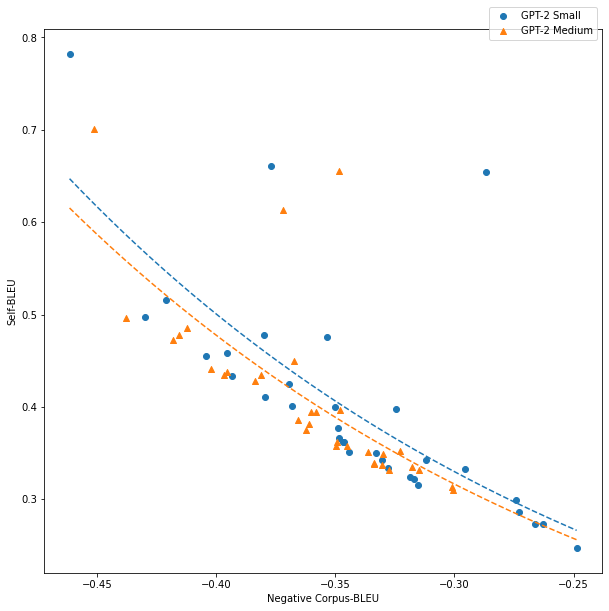}
\caption{Quality vs. Diversity trade-off between GPT-2 Small and GPT-2 Medium models}
\label{SmallVsMedium}
\end{figure}

To see if bigger models are better in the quality/diversity trade-off space, we repeat the experiment using a GPT-2 Medium as a based model. Figure \ref{SmallVsMedium} shows the comparison between Small and Medium, with a log function fitting the data points. Overall, Medium models only perform better than Small models by a narrow margin, as the orange points are closer to the origin of the graph.

\subsection{Consistency}

As discussed in section \ref{Consistency/Commonsense reasoning Lit Review}, we use selection accuracy \citep{li2019dont} as a metric to evaluate a model's ability to demonstrate commonsense reasoning. As we are focusing on story generation, we choose to use the MultiNLI dataset \citep{williams2018nli} and the StoryCloze dataset \citep{mostafazadeh2016corpus} for this experiment.

From the MultiNLI development set, we extract triples of sentences so that for every sentence A, we have a sentence B that contradicts it and a sentence C that entails it. We only consider the triple if the first sentence A has a punctuation at the end of the sentence for the ease of separation. In the end, we have $4,607$ triples of sentences from the MultiNLI development set for our experiment. For every triples, we calculate the perplexity of the second sentence (B or C) with the first sentence A as context. Similar to \citet{li2019dont}, we assume the model will select the sentence with lower perplexity, and we use that to calculate its selection accuracy. Note that this is an unsupervised setup, i.e. we do not fine-tune our models on the MultiNLI dataset. We only use them to compute the perplexity score.

For StoryCloze, we use its development set which consists of $1,570$ stories. We concatenate the first four sentences of each story as the context, and ask the model to select between two different endings. Similar to above, we calculate the perplexity of the two endings given the context to determine the model selection. Results are given in Table \ref{ConsistencyResult}.

\begin{table}[]
\centering
\begin{tabular}{|c||c|c|}
\hline
\textbf{Model}                       & \textbf{MultiNLI} & \textbf{StoryCloze} \\ \hline\hline
GPT-2 Small                          & \textbf{0.65}     & \textbf{0.60}       \\ \hline
MLE GPT-2 Small               & 0.61              & 0.59                \\ \hline
UL GPT-2 Small (1 epoch)   & 0.58              & 0.59                \\ \hline
UL GPT-2 Small (4 epochs)  & 0.54              & 0.59                \\ \hline
GPT-2 Medium                         & \textbf{0.67}     & \textbf{0.67}       \\ \hline
MLE GPT-2 Medium              & 0.63              & 0.66                \\ \hline
UL GPT-2 Medium (1 epoch)  & 0.61              & 0.65                \\ \hline
UL GPT-2 Medium (4 epochs) & 0.57              & 0.64                \\ \hline
\end{tabular}
\caption{Selection accuracy between different models with MultiNLI and StoryCloze dataset}
\label{ConsistencyResult}
\end{table}

It is clear that fine-tuning a model on the Harry Potter corpus decreases its selection accuracy in both dataset, however this is worse in MultiNLI than StoryCloze. One possible explanation is that in StoryCloze we have much longer context, which may help the model in finding the correct ending regardless of the fine-tuning process. 

Overall, UL models perform worse than its MLE counterpart in both dataset. With the MultiNLI dataset, fine-tuning both GPT-2 Small and Medium models using unlikelihood training for 4 epochs decreases their selection accuracy by around 10\%, while this is only around 4\% for maximum likelihood estimate training. This suggests that using unlikelihood training might have a detrimental effect on a model's logical reasoning ability.

Among of all of our trained small models, MLE GPT-2 achieves the best result with a selection accuracy of 0.61 in MultiNLI and 0.59 in StoryCloze. However, this is just slightly better than chance, which might explain why the text produced by the model in Section 3.3 is not consistent.

In all cases, the models which are based off GPT-2 Medium perform considerably better than those are based off GPT-2 Small. This suggests larger models can be superior to smaller models at language understanding.

\subsection{Effect of different domains on consistency}

Training the model on a narrow domain dataset such as Harry Potter series may have a detrimental effect on its selection accuracy, especially with the MultiNLI dataset since it is made up from a variety of domains. To make sure that narrow domain did not cause the drop in selection accuracy with unlikelihood training, we repeat the same experiment but train the models using the WikiText-2 dataset \citep{merity2016pointer}, which we considered as a broader domain dataset compared to the Harry Potter series. The WikiText-2 training set contains 600 Wikipedia articles, with 2,088,628 word tokens in total. From this training set, we randomly select 6,220 sequences of 200 sub-words to match the size of the Harry Potter dataset. We train all of the models using the same environment and hyper-parameters as the previous section.

The final result is given in Table \ref{DomainDifferenceConsistencyResult}. Overall, the new models which were trained using the WikiText-2 performed better than the old ones in selection accuracy in both datasets. However, the trend is still the same: in the MultiNLI selection task, unlikelihood training leads to a lower selection accuracy than the usual maximum likelihood estimate training. This suggests that the drop in selection accuracy when using unlikelihood training has little to do with domain difference.

An interesting thing to note is that when being trained with the WikiText-2 dataset, using more epochs on unlikelihood training actually leads to a slight increase in selection accuracy in both tasks. This further suggests that unlikelihood training is sensitive to the number of training epochs and rather difficult to train.

\begin{table}[]
\centering
\begin{tabular}{|c||c||c|c|}
\hline
\textbf{Corpus}               & \textbf{Model}            & \textbf{MultiNLI} & \textbf{StoryCloze} \\ \hline\hline
\multirow{3}{*}{Harry Potter} & MLE GPT-2 Small           & \textbf{0.61}     & 0.59                \\ \cline{2-4} 
                              & UL GPT-2 Small (1 epoch)  & 0.58              & 0.59                \\ \cline{2-4} 
                              & UL GPT-2 Small (4 epochs) & 0.54              & 0.59                \\ \hline\hline
\multirow{3}{*}{WikiText-2}   & MLE GPT-2 Small           & \textbf{0.64}     & \textbf{0.61}                \\ \cline{2-4} 
                              & UL GPT-2 Small (1 epoch)  & 0.58              & 0.60                \\ \cline{2-4} 
                              & UL GPT-2 Small (4 epochs) & 0.59              & \textbf{0.61}       \\ \hline
\end{tabular}
\caption{Selection accuracy when being trained with different domains}
\label{DomainDifferenceConsistencyResult}
\end{table} 

\lhead{\emph{Multi-task Learning}}
\chapter{Multi-task Learning}

In machine learning, we typically build a model to solve a single well-defined task. We do this by first selecting a metric to measure the models performance for that task, then training the model with the help of a loss function to optimize for that metric. Coming back to the example where we would like to build a machine learning model to categorize images of dogs and cats, the metric here can be the accuracy of predictions, and we can train this model using a binary categorical loss to optimize its parameters for predictions.

However, instead of focusing on that one single task, we can train our machine learning models to solve different tasks simultaneously while still sharing the models parameters across all tasks. This strategy is known as multi-task learning \citep{caruana1997multitask}, and it has been shown to help machine learning models to generalize better on their original task, since they can learn valuable information from a variety of tasks.

Even though multi-task learning has been applied to language models from an early stage in the NLP field \citep{collobert2008unified}, it has grown even more popular with the introduction of BERT \citep{devlin-etal-2019-bert}. BERT, or \textit{\textbf{B}idirectional \textbf{E}ncoder \textbf{R}epresentations from
\textbf{T}ransformers}, is a Transformer-based language model, with the largest variant containing around 340 million parameters. As opposed to GPT-2 - a unidirectional language model, where at every time step the model can only attend to previous words to generate the hidden state for the current word - BERT is bidirectional, meaning that each word can access both of its left (previous) and right (following) context. This bidirectional architecture has been shown to have a much deeper representation of the language context, which allows BERT to obtain new state-of-the-art results on a variety of natural language processing tasks, including question answering and natural language inference \citep{devlin-etal-2019-bert}.

\begin{figure}[t!]
    \centering
    \begin{subfigure}[t]{\textwidth}
        \centering
        \includegraphics[width=12cm]{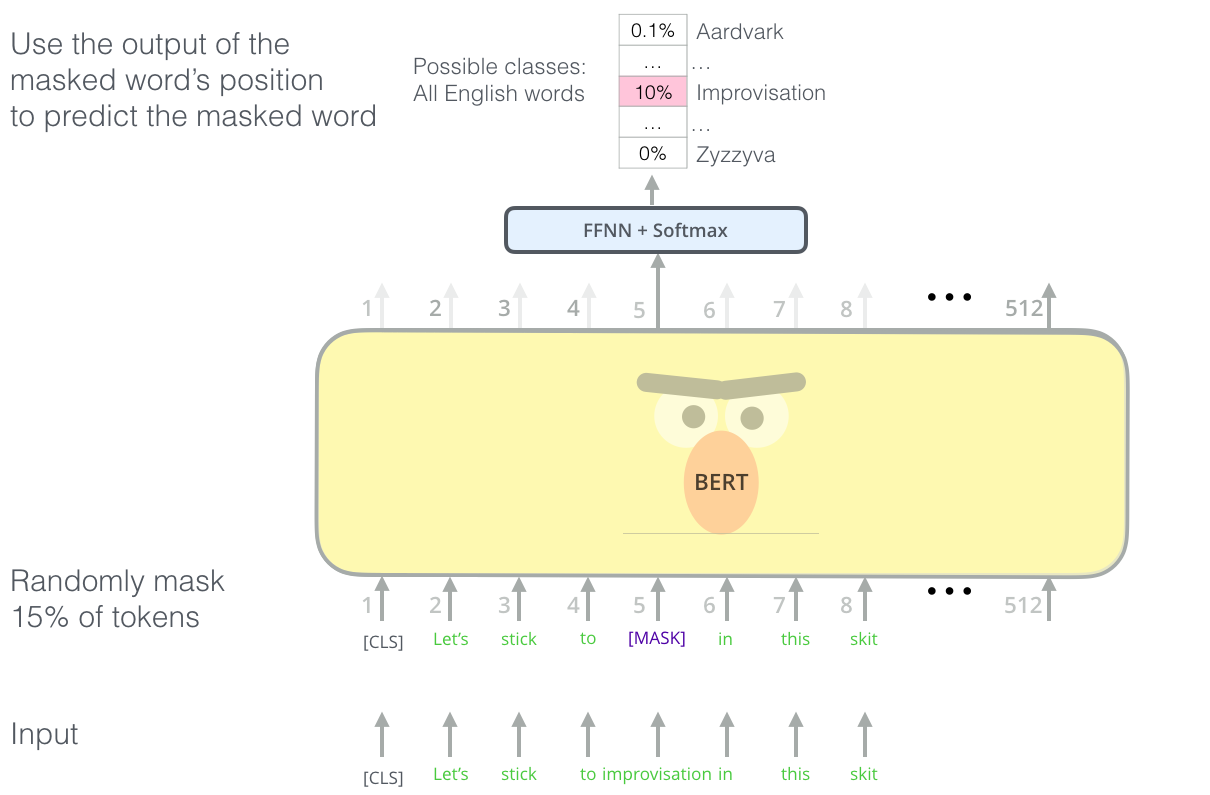}
        \caption{Masked language model}
        \label{BertMLM}
    \end{subfigure}
    \begin{subfigure}[t]{\textwidth}
        \centering
        \includegraphics[width=12cm]{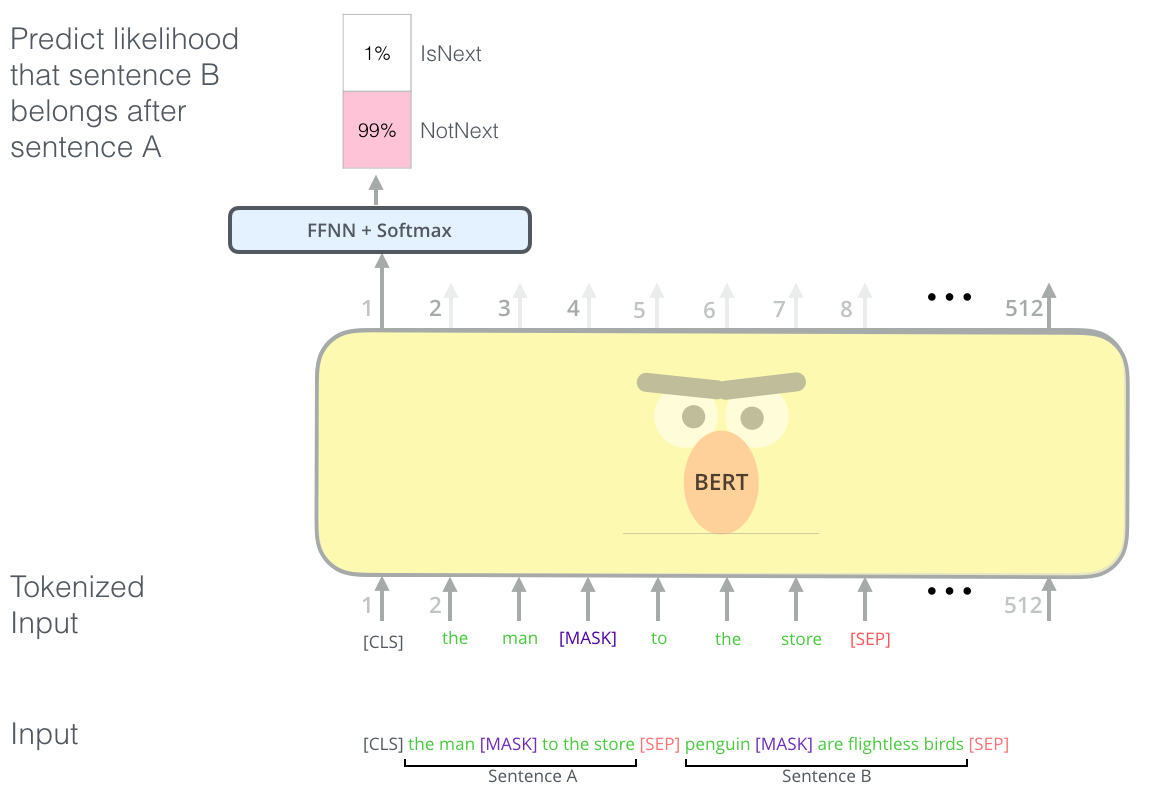}
        \caption{Next sentence prediction}
        \label{BertNSP}
    \end{subfigure}
    \caption{BERT training objectives. Image credit: \citet{alammar2018bert}}
    \label{BertTrainingObjectives}
\end{figure}

BERT obtains its powerful bidirectional representation of words by using two novel training objectives: masked language model and next sentence prediction (Figure \ref{BertTrainingObjectives}). With the masked language model objective, the authors randomly mask 15\% of the words in a training sequence, and ask BERT to correctly identify the masked word (Figure \ref{BertMLM}). With the next sentence prediction objective, the authors first extract pairs of connected sentences from the training corpus, then randomly sample pairs of unrelated sentences. They then give these pairs of sentences to BERT and ask it to predict whether the first sentence is followed by the second sentence in each pair (Figure \ref{BertNSP}).

One downside of BERT is that since each word has access to both of its left and right context, BERT loses the ability of a causal language model to generate continuations, where the model only allows to condition on the context it has seen so far. However, with the success of BERT with multi-task learning, we believe it has the potential to bring causal language models to another level at natural language understanding. In this chapter, we explore different strategies to apply multi-task learning on training causal language models.

\section{Setup}

\subsection{Dataset}

In this experiment, we use the same Harry Potter series from the previous chapter as our training corpus.

\subsection{Training}

We train a new model by fine-tuning the GPT-2 model on the Harry Potter dataset using the original next word prediction (MLE) task with one auxiliary training objective. We end up with 5 different models, which we call NSP GPT-2 (Next Sentence Prediction), SOP GPT-2 (Sentence Order Prediction), TFIDF GPT-2 (Term Frequency - Inverse Document Frequency), POS GPT-2 (Part-of-speech Tag) and DP GPT-2 (Dependency Parsing).

All models are trained with 4 epochs using Adam optimizer with batch size of 5-20 (depending on the memory demand of the training objective) and learning rate of 0.001. In each training batch, we add the losses from all objectives together for back-propagation. The implementation detail of each auxiliary objective is given in the following section.

\subsection{Auxiliary Objectives}


\subsubsection{Sentence-level objective}

To obtain the sentence data for these objectives, we use the \texttt{ntlk} sentence segmenter\footnote{https://www.nltk.org/api/nltk.tokenize.html} to split the Harry Potter dataset into sentences. This results in $92,658$ unique sentences for our experiment.

\subparagraph{Next Sentence Prediction (NSP)}
Inspired by BERT \citep{devlin-etal-2019-bert}, in this task we present our model with two pairs of sentences - one consists of two connected sentences (positive pair) while the other contains two unrelated sentences (negative pair) - and ask the model which one it \textit{prefers}. The preference is calculated using the perplexity score (ppl) for each pair of sentence. We use a \textit{margin ranking loss} function to incentivize the model to give lower perplexity to the positive sentence pair:

$$
\mathcal{L}_{MR}(pos,neg) = \textrm{max}(0, \textrm{ppl}(pos) - \textrm{ppl}(neg) + margin)
$$

Using a number of starting sentences, we obtain the positive training set by simply taking the sentence next to the starting sentences, and the negative training set by randomly sample a sentence from the training corpus. To make it easier for training, we only select 6,220 pairs from each set, which match the number of training sequences for the original next word prediction task. Note that this is done when constructing the training data, thus the same positive and negative sets are used for every training epoch.

\subparagraph{Sentence Order Prediction (SOP)}
Inspired by StructBERT \citep{wang2019structbert}, in this task we give our model a pair of sentences and ask it to predict whether the two sentences are in correct order. The implementation detail is similar to NSP task, except that we obtain the negative training set by simply switching the order of the two sentences in the positive training set.

\subsubsection{Token-level objective}

\subparagraph{TF-IDF}
In this task, we ask our model to predict the TF-IDF score for each token in the training dataset. To obtain the true TF-IDF scores as labels for training, we first split our training corpus into $243$ documents, each one consists of $6400$ tokens, and then calculate the TF-IDF scores based on these documents. In order to produce a TF-IDF score, we add a linear layer on top of the original GPT-2 model. We use a \textit{smooth L1-loss} function for the regression task.

\subparagraph{Part-of-speech Tags (POS)}
In this task, we ask our model to predict the correct part-of-speech tags for each token in the training dataset. We obtain weak labels for training using the \texttt{spacy} part-of-speech tagger\footnote{https://spacy.io/usage/linguistic-features\#pos-tagging}. To make the \texttt{spacy} tokenizer compatible with the GPT-2 tokenizer, we use a similar approach to \citet{devlin-etal-2019-bert} to align the tokens produced by the two tokenizers, where we only assign the POS tag to the first sub-component of a word. For tokens that we could not find a possible alignment, we assign them a special label of X (means \textit{Other}). An example is given in Table \ref{AlignmentStrategy}.

\begin{table}[]
\centering
\begin{tabular}{l||l|l|l|l|l|l|l}
\textbf{spacy tokenizer} & Jane  & \multicolumn{2}{l|}{Doe} & is & a & \multicolumn{2}{l}{musketeer} \\\hline
\textbf{GPT-2 tokenizer}      & Jane  & Do          & \#e       & is & a & musket         & \#eer        \\\hline
\textbf{POS tags}             & I-PER & I-PER       & X         & O  & O & O              & X           
\end{tabular}
\caption{Example of tokenizers alignment strategy}
\label{AlignmentStrategy}
\end{table}

We add a classification head on top of the original GPT-2 model, which uses the last hidden state of each token to predict the POS tag. We use a \textit{cross-entropy loss} function for the classification task.

\begin{figure}[h]
\centering
\includegraphics[width=8cm]{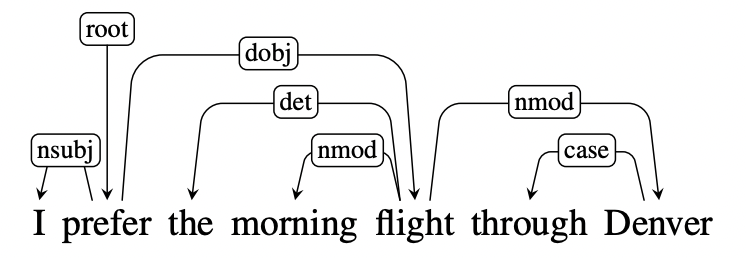}
\caption{Example of dependency parsing \citep{jurafsky2009speech}}
\label{DependencyExample}
\end{figure}

\subparagraph{Dependency Parsing (DP)}
In this task, we ask our model to correctly identify the relationship between certain pair of tokens in the training dataset. Borrowing an example from \citet{jurafsky2009speech}, where we have the sentence \textit{``I prefer the morning flight through Denver"}, we would want our model to correctly predict all dependency labels in this sentence (Figure \ref{DependencyExample}).

Similar to the POS task, we extract all dependency pairs in the training sequences using the \texttt{spacy} dependency parser\footnote{https://spacy.io/usage/linguistic-features\#dependency-parse}, then use the same alignment strategy and cross-entropy loss function for the classification task. Note that we do not perform this classification task for every pair of tokens, but only the ones that we obtain from the \texttt{spacy} dependency parser.

\section{Results}

\subsection{Consistency}

The selection accuracy score with the MultiNLI and StoryCloze dataset for each model is given in Table \ref{LearningObjectivesConsistencyResult}. Overall, all models either perform similar to or better than the MLE model, which has been trained using only the original next word prediction task.

Among all of the five auxiliary training objectives in this experiment, the sentence order prediction task proves to be the most effective one, as the SOP model achieved the best score out of every model, and it got really close to the off-the-shelf GPT-2 selection accuracy score, even though it has been fine-tuned to a specific domain dataset like the Harry Potter series. We also experiment with another model where we combine the two most effective tasks according to the selection accuracy scores given in Table \ref{LearningObjectivesConsistencyResult} (TF-IDF and SOP), however this does not further improve the models performance.

To our disappointment, incorporating syntactical features like part-of-speech tags and word relations does not help the model with the commonsense reasoning task, as we do not achieve any improvements in the selection accuracy score with the POS and DP models compared to the MLE model. This suggests that the GPT-2 model has already learned the necessary syntactical information from its original next word prediction task, and it does not benefit from the help of weakly supervised signals.

\begin{table}[]
\centering
\begin{tabular}{|c||c|c|}
\hline
\textbf{Model}                       & \textbf{MultiNLI} & \textbf{StoryCloze} \\ \hline\hline
GPT-2 (Small)                          & \textbf{0.65}     & 0.60       \\ \hline
MLE GPT-2                          & 0.61     & 0.59       \\ \hline
TFIDF GPT-2               & 0.63              & 0.59                \\ \hline
POS GPT-2   & 0.61              & 0.60                \\ \hline
DP GPT-2  & 0.61              & 0.60                \\ \hline
SOP GPT-2                        & \textbf{0.64}     & 0.59      \\ \hline
NSP GPT-2              & 0.63              & 0.60                \\ \hline
TFIDF+SOP GPT-2  & \textbf{0.64}              & 0.59                \\ \hline
\end{tabular}
\caption{Selection accuracy between different models with MultiNLI and StoryCloze dataset}
\label{LearningObjectivesConsistencyResult}
\end{table}

\subsection{Quality vs. Diversity}

\begin{figure}[h]
\centering
\includegraphics[width=15cm]{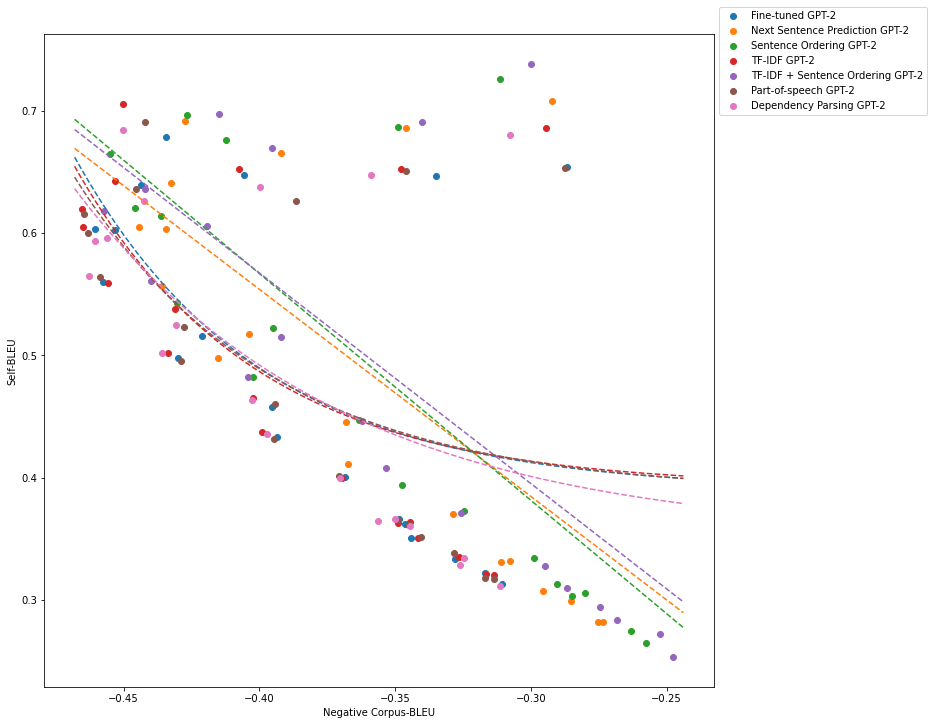}
\caption{Quality vs. Diversity trade-off between different learning objectives}
\label{LearningObjectivesQualityDiversityTradeOff}
\end{figure}

Using Corpus-BLEU and Self-BLEU as the two evaluation metrics for quality and diversity, we plot the quality/diversity trade-off between the models in Figure \ref{LearningObjectivesQualityDiversityTradeOff} and use a log function to fit the data points. Similar to the experiment in Chapter 3, we use Negative Corpus-BLEU instead of Corpus-BLEU, so that the lower is better for both metrics.

Even though NSP, SOP and TF-IDF + SOP are the ones that have achieved the highest selection accuracy scores in the commonsense reasoning task, they did slightly worse at the quality-diversity trade-off space according to the fitting lines in Figure \ref{LearningObjectivesQualityDiversityTradeOff}, which suggests there might be a further trade-off between consistency and quality-diversity. For the other models, it is hard to decide which one is the best model in the quality-diversity trade-off space, since they all have similar performance. Curiously, even though DP curve starts out very similar to MLE, as quality dips it produces the best diversity score (lower right corner, Figure \ref{LearningObjectivesQualityDiversityTradeOff}).

We also give the sequence repetition scores for each model in Table \ref{LearningObjectivesSeqRep4Result}. Overall, the scores are pretty consistent between all models without any outliers, which suggests that by adding these auxiliary training objectives, we do not affect the model in terms of n-gram repetitions in its generation.

\begin{table}[]
\centering
\begin{tabular}{|c||c|c|c|c|c|c|c|}
\hline
\textbf{Model}                       & \textbf{MLE}   & \textbf{TFIDF} & \textbf{POS}   & \textbf{DP}    & \textbf{SOP}   & \textbf{NSP}   & \textbf{TFIDF+SOP} \\ \hline\hline
\textit{Greedy}                      & \textbf{0.584} & \textbf{0.569} & \textbf{0.582} & \textbf{0.590} & \textbf{0.584} & \textbf{0.583} & \textbf{0.619}     \\ \hline
\textit{p=0.2}                       & \textbf{0.474} & \textbf{0.453} & \textbf{0.464} & \textbf{0.478} & \textbf{0.458} & \textbf{0.467} & \textbf{0.508}     \\ \hline
\textit{p=0.4}                       & \textbf{0.203} & \textbf{0.184} & \textbf{0.197} & \textbf{0.204} & \textbf{0.184} & \textbf{0.188} & \textbf{0.219}     \\ \hline
\textit{p=0.6}                       & 0.054          & 0.049          & 0.049          & 0.054          & 0.050          & 0.054          & 0.059              \\ \hline
\textit{p=0.8}                       & 0.017          & 0.017          & 0.017          & 0.018          & 0.018          & 0.018          & 0.017              \\ \hline
\textit{p=0.9}                       & 0.010          & 0.011          & 0.010          & 0.011          & 0.010          & 0.012          & 0.010              \\ \hline
\textit{k=2}   & \textbf{0.213} & \textbf{0.198} & \textbf{0.209} & \textbf{0.211} & \textbf{0.210} & \textbf{0.209} & \textbf{0.234}     \\ \hline
\textit{k=10}  & 0.024          & 0.024          & 0.024          & 0.026          & 0.026          & 0.026          & 0.028              \\ \hline
\textit{k=50}  & 0.010          & 0.010          & 0.010          & 0.011          & 0.011          & 0.011          & 0.011              \\ \hline
\textit{k=100} & 0.009          & 0.009          & 0.008          & 0.008          & 0.009          & 0.009          & 0.009              \\ \hline
\textit{k=200} & 0.007          & 0.007          & 0.007          & 0.008          & 0.008          & 0.008          & 0.008              \\ \hline\hline
\textit{Human Text*} & \multicolumn{7}{c|}{0.007}                        \\ \hline
\end{tabular}
\caption{Sequence repetition scores between different learning objectives and decoding methods. *Human text sequence repetition is computed using the training dataset}
\label{LearningObjectivesSeqRep4Result}
\end{table}

Overall, adding these auxiliary objectives seem to help improve logical consistency, while they do not harm the model in terms of quality-diversity trade-off and sequence repetition. This suggests that there might be an incentive for including these auxiliary objectives when training/fine-tuning a language model.

\lhead{\emph{Conclusion}}
\chapter{Conclusion}

\section{Overview}

In this work, we have provided necessary backgrounds to understand language models and the problem that they have in open-ended text generation task, which are (i) neural language degeneration and (ii) the lack of consistent evaluation metrics to measure their performance. It is important that we evaluate language models in all dimensions of open-ended text generation - quality, diversity and consistency.

We have conducted an experiment to search for the best metric to use when evaluating language models on open-ended text generation task. We have proposed an evaluation pipeline using these metrics and apply it to compare the performance between different stochastic decoding methods and unlikelihood training.

Finally, we have carried out experiments with multi-task learning to see if it can help language models to get better at open-ended text generation. Using the evaluation pipeline above, we evaluated the efficacy of different auxiliary training objectives in open-ended text generation task.

\section{Contributions}

\myparagraph{A practical pipeline to evaluate language models on open-ended text generation task}

When evaluating language models on open-ended text generation task, we have found that Corpus-BLEU is the best metric to evaluate the quality of generated text due to its similarity with human judgement. As for diversity, Self-BLEU appears to be the best metric to use thanks to its simplicity to calculate. To evaluate the consistency of the generated text, using selection accuracy on the MultiNLI dataset is good enough for most cases. For specific task such as story generation, other dataset can be considered (e.g. StoryCloze).

\myparagraph{A direct comparison between unlikelihood training and stochastic decoding methods}

To the best of our knowledge, there has not been any works that \textit{quantitatively} compare unlikelihood training with stochastic decoding methods. Using our proposed evaluation pipeline, we found out that there was no clear difference between unlikelihood training and maximum likelihood estimate training with stochastic decoding methods in the quality-diversity trade-off space. However, unlikelihood training might have a negative effect on the ability of a language model to truly grasp the gist of the language, as it worsens the model performance in commonsense reasoning task.

\myparagraph{An insight on how multi-task learning can lead to better machine generation}

We found out that by adding certain auxiliary training objectives along with the maximum estimate likelihood objective when fine-tuning GPT-2 models, they achieved a much better score in commonsense reasoning task, while still maintain their performance in the quality-diversity trade-off space. This suggests that multi-task learning might help a language model to truly understand the language, which in turn leads to better generation.

\section{Future Work}

\myparagraph{Incorporate human evaluation to verify the correctness of the evaluation metrics}

Even though our experiment with evaluation metrics agree with \citet{zhang2020trading}'s finding of \textit{the likelihood trap}, further experiment with the help of human evaluation should be carried out to verify that the metrics work as expected. To the best of our knowledge, human judgement has only been used to evaluate the quality of generation, since it might be hard for one to assess the diversity of machine-generated text \citep{hashimoto2019unifying}. Therefore, it might be worth to investigate how we can leverage human evaluation to confirm the correctness of diversity metrics as well. 

\myparagraph{Further experiments on different ways to tweak language models training objective} 

In this work, we only select unlikelihood training \citep{welleck2019neural} as a representative for the training objective strategy when making comparison with stochastic decoding methods. For future work, it would be useful to look at different strategies to tweak language models training objective as well, such as scheduled sampling \citep{bengio2015scheduled}, GAN \citep{yu2017seqgan, guo2017long} and entmax loss \citep{martins2020sparse}.

\myparagraph{Experiment with more auxiliary training objectives}


In this work, we only focused on syntactic supervision objectives. For future work, it might be worth to carry out experiment with training objectives that contain more semantics-oriented information, such as word senses or topics classification.

\myparagraph{Experiment with larger language models}

Due to the limitation of the available computing resource, we could only perform our experiment on GPT-2 Small and Medium models. If one has access to larger language models, it might be worth to repeat the same experiment to see how much better do larger language models get at open-ended text generation. 







\addtocontents{toc}{\vspace{2em}}  
\backmatter

\label{Bibliography}
\lhead{\emph{Bibliography}}  
\bibliography{Bibliography}  

\end{document}